\documentclass{article}

\usepackage{microtype}
\usepackage{graphicx}
\usepackage{subcaption}
\usepackage{booktabs} % for professional tables

\usepackage{hyperref}

\usepackage[preprint]{icml2026}

\usepackage{amsmath}
\usepackage{amssymb}
\usepackage{mathtools}
\usepackage{amsthm}

\usepackage[capitalize,noabbrev]{cleveref}

\theoremstyle{plain}

\theoremstyle{definition}

\theoremstyle{remark}

\usepackage[textsize=tiny]{todonotes}

\usepackage[utf8]{inputenc} % allow utf-8 input
\usepackage[T1]{fontenc}    % use 8-bit T1 fonts
\usepackage{hyperref}       % hyperlinks
\usepackage{url}            % simple URL typesetting
\usepackage{booktabs}       % professional-quality tables
\usepackage{amsfonts}       % blackboard math symbols
\usepackage{nicefrac}       % compact symbols for 1/2, etc.
\usepackage{microtype}      % microtypography
\usepackage{xcolor}         % colors

\usepackage{hyperref}
\usepackage{dsfont}
\usepackage{url}
\usepackage{natbib}
\usepackage{amssymb, amsmath,amsthm}
\usepackage{amsfonts}
\usepackage{algorithm}
\usepackage{algorithmic}
\usepackage{paralist}
\usepackage{multirow}
\usepackage{times}
\usepackage{wrapfig}

\usepackage{url,enumerate}
\usepackage{color,xcolor}
\usepackage{makeidx}  % allows for indexgeneration
\usepackage{amsmath,amssymb}
\usepackage{mathtools}
\usepackage[small, compact]{titlesec}
\usepackage{xspace}
\usepackage{epstopdf}
\usepackage{mathrsfs}
\usepackage{times}
\usepackage{enumerate}
\usepackage{color}
\usepackage{graphicx,epsfig}
\usepackage{amsmath,amssymb,xspace}
\usepackage{url}
\usepackage{bm}
\usepackage{bbm}
\usepackage{upgreek}
\usepackage{cleveref}
\usepackage{multirow}
\usepackage{ulem}
\usepackage{cancel}
\usepackage{empheq}
\usepackage{pifont}
\usepackage{subcaption}
\usepackage{amsmath}
\usepackage{amssymb}
\usepackage{mathtools}
\usepackage{amsthm}

\theoremstyle{plain}
\newcommand{\cmt}[1]{}

\newcommand{\ours}{{DGPFM}\xspace}

\icmltitlerunning{Deep Gaussian Processes for Functional Maps}

\begin{document}

\twocolumn[
  \icmltitle{Deep Gaussian Processes for Functional Maps}



  \begin{icmlauthorlist}
    \icmlauthor{Matthew Lowery}{ksc}
    \icmlauthor{Zhitong Xu}{ksc}
    \icmlauthor{Da Long}{ksc}
    \icmlauthor{Keyan Chen}{ksc}
    \icmlauthor{Daniel S. Johnson}{ksc}
    \icmlauthor{Yang Bai}{hk}
    \icmlauthor{Varun Shankar}{ksc}
    \icmlauthor{Shandian Zhe}{ksc}
  \end{icmlauthorlist}

  \icmlaffiliation{ksc}{Kahlert School of Computing, University of Utah}
  \icmlaffiliation{hk}{Department of Health and Kinesiology, University of Utah}

  \icmlcorrespondingauthor{Matthew Lowery, Varun Shankar, and Shandian Zhe}{\{mlowery, shankar, zhe\}@cs.utah.edu}
  \icmlcorrespondingauthor{Zhitong Xu, Da Long, Keyan Chen, and Daniel S. Johnson}{\{u1502956, da.long, u1466725, d.johnson\}@utah.edu}

  \icmlkeywords{Machine Learning, ICML}

  \vskip 0.3in
]



\printAffiliationsAndNotice{}  % no special notice (required even if empty)

% Math commands by Thomas Minka
\newcommand{\var}{{\rm var}}
\newcommand{\Tr}{^{\rm T}}
\newcommand{\vtrans}[2]{{#1}^{(#2)}}
\newcommand{\kron}{\otimes}
\newcommand{\schur}[2]{({#1} | {#2})}
\newcommand{\schurdet}[2]{\left| ({#1} | {#2}) \right|}
\newcommand{\had}{\circ}
\newcommand{\diag}{{\rm diag}}
\newcommand{\invdiag}{\diag^{-1}}
\newcommand{\rank}{{\rm rank}}
\newcommand{\expt}[1]{\langle #1 \rangle}
\newcommand{\whalpha}{\widehat{\alpha}}
% careful: ``null'' is already a latex command
\newcommand{\nullsp}{{\rm null}}
\newcommand{\tr}{{\rm tr}}
\renewcommand{\vec}{{\rm vec}}
\newcommand{\vech}{{\rm vech}}
\renewcommand{\det}[1]{\left| #1 \right|}
\newcommand{\pdet}[1]{\left| #1 \right|_{+}}
\newcommand{\pinv}[1]{#1^{+}}
\newcommand{\erf}{{\rm erf}}
\newcommand{\hypergeom}[2]{{}_{#1}F_{#2}}
\newcommand{\mcal}[1]{\mathcal{#1}}
\newcommand{\gp}{\mathcal{GP}}
% boldface characters
\renewcommand{\a}{{\bf a}}
\renewcommand{\b}{{\bf b}}
\renewcommand{\c}{{\bf c}}
\renewcommand{\d}{{\rm d}}  % for derivatives
\newcommand{\e}{{\bf e}}
\newcommand{\f}{{\bf f}}
\newcommand{\g}{{\bf g}}
\newcommand{\h}{{\bf h}}
\newcommand{\bi}{{\bf i}}
\newcommand{\bj}{{\bf j}}
\newcommand{\bK}{{\bf K}}
%\newcommand{\k}{{\bf k}}
% in Latex2e this must be renewcommand
\renewcommand{\k}{{\bf k}}
\newcommand{\m}{{\bf m}}
\newcommand{\mhat}{{\overline{m}}}
\newcommand{\tm}{{\tilde{m}}}
\newcommand{\n}{{\bf n}}
\renewcommand{\o}{{\bf o}}
\newcommand{\p}{{\bf p}}
\newcommand{\q}{{\bf q}}
\renewcommand{\r}{{\bf r}}
\newcommand{\s}{{\bf s}}
\renewcommand{\t}{{\bf t}}
\renewcommand{\u}{{\bf u}}
\renewcommand{\v}{{\bf v}}
\newcommand{\w}{{\bf w}}
\newcommand{\x}{{\bf x}}
\newcommand{\y}{{\bf y}}
\newcommand{\z}{{\bf z}}
\newcommand{\bl}{{\bf l}}
\newcommand{\A}{{\bf A}}
\newcommand{\B}{{\bf B}}
\newcommand{\C}{{\bf C}}
\newcommand{\D}{{\bf D}}
\newcommand{\Dcal}{\mathcal{D}}
\newcommand{\Ocal}{\mathcal{O}}
\newcommand{\E}{{\bf E}}
\newcommand{\F}{{\bf F}}
\newcommand{\G}{{\bf G}}
\newcommand{\Gcal}{{\mathcal{G}}}
\renewcommand{\H}{{\bf H}}
\newcommand{\I}{{\bf I}}
\newcommand{\J}{{\bf J}}
\newcommand{\K}{{\bf K}}
\renewcommand{\L}{{\bf L}}
\newcommand{\Lcal}{{\mathcal{L}}}
\newcommand{\M}{{\bf M}}
\newcommand{\Mcal}{{\mathcal{M}}}
\newcommand{\Ecal}{{\mathcal{E}}}
\newcommand{\N}{\mathcal{N}}  % for normal density
\newcommand{\TN}{\mathcal{TN}}  % for normal density
\newcommand{\MN}{\mathcal{MN}}
\newcommand{\bupeta}{\boldsymbol{\upeta}}
\newcommand{\kl}{{\text{KL}}}
\renewcommand{\O}{{\bf O}}
\renewcommand{\P}{{\bf P}}
\newcommand{\Q}{{\bf Q}}
\newcommand{\R}{{\bf R}}
\renewcommand{\S}{{\bf S}}
\newcommand{\Scal}{{\mathcal{S}}}
\newcommand{\Bcal}{{\mathcal{B}}}
\newcommand{\Pcal}{{\mathcal{P}}}
\newcommand{\T}{{\bf T}}
\newcommand{\Tcal}{{\mathcal{T}}}
\newcommand{\U}{{\bf U}}
\newcommand{\Ucal}{{\mathcal{U}}}
\newcommand{\tU}{{\tilde{\U}}}
\newcommand{\tUcal}{{\tilde{\Ucal}}}
\newcommand{\V}{{\bf V}}
\newcommand{\W}{{\bf W}}
\newcommand{\Wcal}{{\mathcal{W}}}
\newcommand{\Vcal}{{\mathcal{V}}}
\newcommand{\X}{{\bf X}}
\newcommand{\Xcal}{{\mathcal{X}}}
\newcommand{\Acal}{{\mathcal{A}}}
\newcommand{\Y}{{\bf Y}}
\newcommand{\Ycal}{{\mathcal{Y}}}
\newcommand{\Z}{{\bf Z}}
\newcommand{\Zcal}{{\mathcal{Z}}}
\newcommand{\Hcal}{{\mathcal{H}}}
\newcommand{\Fcal}{{\mathcal{F}}}
\newcommand{\whL}{{\widehat{\Lcal}}}
\newcommand{\whJ}{{\widehat{J}}}

% this is for latex 2.09
% unfortunately, the result is slanted - use Latex2e instead
%\newcommand{\bfLambda}{\mbox{\boldmath$\Lambda$}}
% this is for Latex2e
\newcommand{\bfLambda}{\boldsymbol{\Lambda}}

% Yuan Qi's boldsymbol
\newcommand{\bsigma}{\boldsymbol{\sigma}}
\newcommand{\balpha}{\boldsymbol{\alpha}}
\newcommand{\bpsi}{\boldsymbol{\psi}}
\newcommand{\bphi}{\boldsymbol{\phi}}
\newcommand{\bPhi}{\boldsymbol{\Phi}}
\newcommand{\cov}{{\text{cov}}}

\newcommand{\bbeta}{\boldsymbol{\beta}}
\newcommand{\bepsi}{\boldsymbol{\epsilon}}
\newcommand{\boldeta}{\boldsymbol{\eta}}
\newcommand{\btau}{\boldsymbol{\tau}}
\newcommand{\bvarphi}{\boldsymbol{\varphi}}
\newcommand{\bzeta}{\boldsymbol{\zeta}}

\newcommand{\blambda}{\boldsymbol{\lambda}}
\newcommand{\bLambda}{\mathbf{\Lambda}}

\newcommand{\btheta}{{\boldsymbol{\theta}}}
\newcommand{\bTheta}{\boldsymbol{\Theta}}
\newcommand{\bpi}{\boldsymbol{\pi}}
\newcommand{\bxi}{\boldsymbol{\xi}}
\newcommand{\bSigma}{\boldsymbol{\Sigma}}
\newcommand{\bPi}{\boldsymbol{\Pi}}
\newcommand{\bOmega}{\boldsymbol{\Omega}}
\newcommand{\brho}{\boldsymbol{\rho}}

\newcommand{\bgamma}{\boldsymbol{\gamma}}
\newcommand{\bGamma}{\boldsymbol{\Gamma}}
\newcommand{\bUpsilon}{\boldsymbol{\Upsilon}}
\newcommand{\barZ}{\bar{Z}}
\newcommand{\barz}{\bar{z}}
\newcommand{\whatR}{\widehat{R}}

\newcommand{\bmu}{\boldsymbol{\mu}}
\newcommand{\1}{{\bf 1}}
\newcommand{\0}{{\bf 0}}

\newcommand{\bs}{\backslash}
\newcommand{\ben}{\begin{enumerate}}
\newcommand{\een}{\end{enumerate}}

 \newcommand{\notS}{{\backslash S}}
 \newcommand{\nots}{{\backslash s}}
 \newcommand{\noti}{{\backslash i}}
 \newcommand{\notj}{{\backslash j}}
 \newcommand{\nott}{\backslash t}
 \newcommand{\notone}{{\backslash 1}}
 \newcommand{\nottp}{\backslash t+1}

\newcommand{\notk}{{^{\backslash k}}}
\newcommand{\notij}{{^{\backslash i,j}}}
\newcommand{\notg}{{^{\backslash g}}}
\newcommand{\wnoti}{{_{\w}^{\backslash i}}}
\newcommand{\wnotg}{{_{\w}^{\backslash g}}}
\newcommand{\vnotij}{{_{\v}^{\backslash i,j}}}
\newcommand{\vnotg}{{_{\v}^{\backslash g}}}
\newcommand{\half}{\frac{1}{2}}
\newcommand{\msgb}{m_{t \leftarrow t+1}}
\newcommand{\msgf}{m_{t \rightarrow t+1}}
\newcommand{\msgfp}{m_{t-1 \rightarrow t}}

\newcommand{\proj}[1]{{\rm proj}\negmedspace\left[#1\right]}
\newcommand{\argmin}{\operatornamewithlimits{argmin}}
\newcommand{\argmax}{\operatornamewithlimits{argmax}}

\newcommand{\dif}{\mathrm{d}}
\newcommand{\abs}[1]{\lvert#1\rvert}
\newcommand{\norm}[1]{\lVert#1\rVert}

\newcommand{\ie}{{\textit{i.e.,}}\xspace}
\newcommand{\etc}{{\textit{etc}.}\xspace}
\newcommand{\eg}{{{\textit{e.g.},}}\xspace}
\newcommand{\EE}{\mathbb{E}}
\newcommand{\HH}{\mathbb{H}}
\newcommand{\sbr}[1]{\left[#1\right]}
\newcommand{\rbr}[1]{\left(#1\right)}
\newcommand{\zhe}[1]{{\textcolor{blue}{#1}}}
\newcommand{\Vtr}{\mathrm{Vec}}
\newcommand{\tlam}{{\tilde{\lambda}}}
\newcommand{\tp}{{\widetilde{p}}}
\newcommand{\tmu}{{\widetilde{\mu}}}
\newcommand{\tv}{{\widetilde{v}}}
\newcommand{\talpha}{{\widetilde{\alpha}}}
\newcommand{\tomega}{{\widetilde{\omega}}}
\newcommand{\bkh}{{\backslash}}
\newcommand{\whmu}{\widehat{\bmu}}
\newcommand{\whV}{\widehat{\V}}
\newcommand{\ol}[1]{{\overline{#1}}}

\newcommand{\YM}[1]{\textcolor{blue}{\small {\sf YM: #1}}}

\begin{abstract}
%Learning mappings between functional spaces, also known as function-on-function regression, plays a crucial role in functional data analysis and has broad applications, \eg spatiotemporal forecasting, curve prediction, and climate modeling.	Existing approaches fall short of capturing complex  nonlinearities and/or lack  reliable uncertainty quantification under noisy, sparse, and irregularly sampled data.   To address these issues, we propose Deep Gaussian Processes for Functional Maps (\ours).  Our method designs a sequence of GP-based linear and nonlinear transformations in function space, leveraging integral transforms of kernels, GP conditional means, and nonlinear activations sampled from GPs. A key insight simplifies implementation: under fixed  locations, discrete approximations of kernel integral transforms collapse into direct functional integral transforms, enabling flexible incorporation of various integral transform designs. To achieve scalable probabilistic inference, we use inducing points and whitening transformations to develop a variational learning algorithm. Empirical results on real-world and synthetic benchmark datasets demonstrate that the advantage of \ours in both predictive performance and uncertainty calibration.
Learning mappings between functional spaces, also known as function-on-function regression, is a fundamental problem in functional data analysis with broad applications, including spatiotemporal forecasting, curve prediction, and climate modeling. Existing approaches often struggle to capture complex nonlinear relationships and/or provide reliable uncertainty quantification when data are noisy, sparse, or irregularly sampled. To address these challenges, we propose Deep Gaussian Processes for Functional Maps (\ours). Our method constructs a sequence of GP-based linear and nonlinear transformations directly in function space, leveraging kernel integral transforms, GP conditional means, and nonlinear activations sampled from Gaussian processes. A key insight enables a simplified and flexible implementation: under fixed evaluation locations, discrete approximations of kernel integral transforms reduce to direct functional integral transforms, allowing seamless integration of diverse transform designs. To support scalable probabilistic inference, we adopt inducing points and whitening transformations within a variational learning framework. Empirical results on both real-world and synthetic benchmark datasets demonstrate the advantages of \ours in terms of predictive accuracy and uncertainty calibration.

\end{abstract}
%function on function maps --> linear method, recent works in neural operators is for PDE operator learning, point estimate, etc. --> we propose deep GP based models, contributions --> experiments
\section{Introduction}
%Function-on-function regression~\citep{ramsay1991some,morris2015functional} extends standard regression into functional spaces, where both input and output variables are functions --- objects that are infinite-dimensional in nature. It serves as a fundamental tool in functional data analysis~\citep{ramsay2002applied} and has found widespread applications such as temporal, spatiotemporal, and curve prediction in econometrics~\citep{rust2022improving}, brain imaging~\citep{wang2014regularized,wang2013regularized}, energy and utility consumption forecasting~\citep{fumo2015regression}, and weather and climate modeling~\citep{holmstrom2016machine,masselot2018new}.

Function-on-function regression~\citep{ramsay1991some,morris2015functional} extends standard regression to functional spaces, where both inputs and outputs are functions --- objects that are infinite-dimensional in nature. It is a fundamental tool in functional data analysis~\citep{ramsay2002applied} and has found widespread applications in temporal and spatiotemporal prediction, curve forecasting, and related tasks across econometrics~\citep{rust2022improving}, brain imaging~\citep{wang2014regularized,wang2013regularized}, energy and utility consumption forecasting~\citep{fumo2015regression}, and weather and climate modeling~\citep{holmstrom2016machine,masselot2018new}.

%Despite the success of existing methods, most of them have focused on functional linear regression~\citep{yao2005functional,manrique2016functional}, which predicts the output function by integrating the input function against a (parameterized) regression function. This effectively performs a linear transformation in an infinite-dimensional space. While intuitive and elegant, simple linear transformations often fall short in capturing the complex nonlinear relationships arising in real-world applications. In scientific machine learning, neural network models --- often referred to as neural operators~\citep{li2020fourier,azizzadenesheli2024neural} --- have been proposed to learn the operators of partial differential equations (PDEs), which can be viewed as a special case of function-on-function regression. Although successful, these methods typically rely on high-quality, noise-free, regularly sampled data generated from numerical simulations. Moreover, they focus primarily on point estimation and lack mechanisms for uncertainty quantification. In practical settings,
%however, data are often noisy, sparse, and irregularly sampled, necessitating models that are not only highly expressive but also robust and capable of providing well-calibrated uncertainty estimates.
Despite the success of existing approaches, most prior work has focused on functional linear regression~\citep{yao2005functional,manrique2016functional}, where the output function is obtained by integrating the input function against a (parameterized) regression function. This corresponds to a linear transformation in an infinite-dimensional space. While intuitive and elegant, such linear models often fail to capture the complex nonlinear relationships that arise in real-world applications. In scientific machine learning, neural network–based models --- commonly referred to as neural operators~\citep{li2020fourier,azizzadenesheli2024neural} --- have been proposed to learn operators associated with partial differential equations (PDEs), which can be viewed as a special case of function-on-function regression. Although successful, these methods typically rely on high-quality, noise-free, and regularly sampled data generated from numerical simulations, and they primarily focus on point estimation without providing principled uncertainty quantification. In contrast, real-world data are often noisy, sparse, and irregularly sampled, necessitating models that are both highly expressive and capable of producing well-calibrated uncertainty estimates.

 To address these challenges, we propose \ours, a deep Gaussian process model for functional maps. \ours is flexible enough to capture complex, highly nonlinear relationships and, importantly, enables principled probabilistic inference with reliable uncertainty quantification. We model the input function as a Gaussian process (GP) and represent the mapping from inputs to outputs as a sequence of GP-based linear and nonlinear transformations in function space. The linear transformation is implemented via a kernel integral transform to obtain cross-covariances, followed by GP interpolation, while the nonlinear transformation is realized through a GP activation function. A key insight of our design is that, under fixed evaluation locations, any discrete approximation of the kernel integral transform leads to cancellation of intermediate covariance and cross-covariance matrices during GP interpolation. As a result, the integral transform can be applied directly to discretized functions, eliminating the need to track complex kernel structures across layers and substantially simplifying implementation. This insight allows our framework to flexibly incorporate arbitrary discrete integral transform designs without explicitly computing the resulting covariance functions. In particular, we consider simple dimension-wise integral transforms, implemented either via one-dimensional quadrature rules or via the convolution theorem and Fourier transforms, inspired by the neural operator literature. To enable scalable training and probabilistic inference, we introduce inducing points for each GP activation and apply a whitening transformation to construct a variational posterior, leading to an efficient stochastic variational inference algorithm.

%For evaluation, we assess \ours on three synthetic datasets and three real-world applications. \ours nearly always achieves the best prediction accuracy in terms of normalized root mean square error (NRMSE). More importantly, it outperforms alternative methods in test log-likelihood scores, demonstrating substantially better uncertainty calibration than Bayesian version of neural operators trained with popular methods such as Stochastic Gradient Langevin Dynamics and Monte Carlo dropout. Visualizations of prediction examples further confirm that \ours produces reliable and well-calibrated uncertainty estimates.
We evaluate \ours on three synthetic benchmarks and three real-world applications. \ours consistently achieves strong predictive performance in terms of normalized root mean square error (NRMSE) and, more importantly, significantly outperforms alternative methods in test log-likelihood. These results demonstrate substantially improved uncertainty calibration compared to Bayesian variants of neural operators trained using popular approaches such as stochastic gradient Langevin dynamics and Monte Carlo dropout. Qualitative visualizations further confirm that \ours produces reliable and well-calibrated uncertainty estimates.
%Gaussian processes & linear functional regression 
\section{Background}
\noindent\textbf{Functional Linear Regression.} Function-on-function regression aims to estimate the mapping between two functional spaces $\Fcal_1$ and $\Fcal_2$. Given an input function  $f(\cdot) \in \Fcal_1$ and output function $u(\cdot)  \in \Fcal_2$, functional linear regression (FLR)~\citep{yao2005functional} introduces a linear mapping:  $u(\x) = \int w(\x, \x') f(\x') \d \x'$, 
%\begin{align}
%	u(\x) = \int w(\x, \x') f(\x') \d \x', \label{eq:flr}
%\end{align}
where $w(\x, \x')$ is the coefficient function, extending standard  linear regression to an infinite dimensional space.\cmt{ --- treating all the  values of  $f$ and $u$ as predictors and responses, respectively.} To make estimation tractable, FLR typically employs basis function expansions to represent $f$, $u$, and $w$ in finite-dimensional forms, \eg $u(\x) = \sum_{k=1}^K c_k \phi_k(\x)$, $f(\x') = \sum_{l=1}^L \alpha_l\psi_l(\x')$, and $w(\x, \x') = \sum_{k=1}^K \sum_{l=1}^L \omega_{kl} \phi_k(\x)\psi_l(\x')$, where $\{\phi_k\}$ and $\{\psi_l\}$ are basis functions, and $\omega_{kl}$ are the coefficients of the regression surface.  The model can then be expressed as multi-variate linear regression. \cmt{, 
$\c = \W \z$ where $\c = [c_1, \cdots, \c_K]$, $\W =[\omega_{ij}]$, and $\z = [z_1, \ldots, z_L]$ with each $z_l = \sum_j \alpha_j \int \psi_j(\x') \psi_l(\x')\d\x'$.} Commonly used bases  include B-spines, Fourier bases, and others. 

\noindent\textbf{Gaussian Processes (GPs).}  Gaussian processes offer a powerful probabilistic  framework for function estimation. Let $f: \mathbb{R}^d \rightarrow \mathbb{R}$ denote the target function.  A GP places a prior over $f$ such that:  $f(\cdot) \sim \gp(m(\cdot), \cov(\cdot,\cdot))$,  where $m(\cdot)$ is the mean function and $\cov(\cdot, \cdot)$ is the covariance function, often specified as a kernel function $k(\x, \x')$.  In practice,  $m$ is usually set to zero. %A popular kernel choice is the square exponential (SE) kernel, $k(\x, \x') = \exp\left(- \|\x - \x'\|^2/\rho\right)$ where $\rho>0$ is the length-scale parameter. 
Given input locations $\X = [\x_1, \ldots, \x_N]^\top$, the corresponding function values $\f = [f(\x_1), \ldots, f(\x_N)]$, follow a multi-variate Gaussian distribution, $p(\f) = \N(\f|\0, \K)$ where $[\K]_{ij} = \cov(f(\x_i), f(\x_j)) = k(\x_i, \x_j)$. This projection is fundamental to GP inference. 
%Under the GP prior, the function values at any finite set of  input locations, $\f = [f(\x_1), \ldots, f(\x_N)]$, follow a multi-variate Gaussian distribution, $p(\f) = \N(\f|\0, \K)$ where $[\K]_{ij} = \cov(f(\x_i), f(\x_j)) = k(\x_i, \x_j)$. This is called a GP projection.
 Suppose $\f$ is known,  and we want to predict the function value at a new location $\x$. Since $\f$ and $f(\x)$ also follow a multi-variate Gaussian distribution, we immediately obtain a conditional Gaussian as the predictive distribution, $p(f(\x) | \f) = \N\left(f(\x) | \mu(\x), \sigma^2(\x)\right)$, where the conditional mean gives an interpolation estimate, $\mu(\x) = \cov(f(\x), \f) \K^{-1} \f$, 
%\begin{align}
%	\mu(\x) = \cov(f(\x), \f) \K^{-1} \f,  \label{eq:gp-int}
%\end{align}
and the conditional variance $\sigma^2(\x) = \cov(f(\x), f(\x)) -  \cov(f(\x), \f) \K^{-1}  \cov(\f, f(\x))$ quantifies the prediction uncertainty, $\cov(f(\x), \f) = k(\x, \X) = [k(\x, \x_1), \ldots, k(\x, \x_N)]$ and $\X = [\x_1, \ldots, \x_N]^\top$.

%motiation again, linear comonents, simplication and then overall archiecture later 
\section{Model}
We now introduce \ours, our deep GP model for learning mappings between functions. 
Given an input function $f(\cdot)$ and an output function $u(\cdot)$, we model $f$ as a GP, and construct the mapping from $f$ to $u$ through a sequence of intermediate conditional GP layers. These layers implement successive linear and nonlinear transformations directly in function space, enabling flexible and expressive function-to-function modeling. An overview of our approach is illustrated in Appendix Figure~\ref{fig:model-arch}.

\subsection{GP-based Linear and Nonlinear Transformation}\label{sect:gp-transform}
Let $C$ denote the number of GPs in each layer. 
 At layer $l$, we denote the $i$-th GP  ($1 \le i \le C$) by $h_{l,i}(\cdot)$, with associated covariance function $\kappa_{l,i}(\cdot, \cdot)$.  These GPs are defined conditionally on the latent functions of the preceding layer; when marginalized over previous layers, they do not, in general, remain GPs.  To perform a \textit{linear} transformation in function space, we introduce a coefficient function $w_{l}(\cdot, \cdot)$ and model 
\begin{align}
	h_{l+1, i}(\x) = \int w_{l}(\x, \x')h_{l,i}(\x') \d \x',  \label{eq:int-transform}
\end{align}
which is closely related to FLR. {Unlike FLR, where $w_l$ is typically parameterized by a fixed functional form, we place a GP prior on the coefficient function, $w_l \sim \gp$, to enhance modeling flexibility.} {Conditioning on a realization of $w_l$, } the integral transform of a GP remains a GP. Consequently, $h_{l+1,i}(\cdot)$ is a GP with  covariance and cross-covariance functions\footnote{The detail derivation is provided in Appendix section~\ref{sect:cross-cov}.}  given by 
\begin{align}
	&\kappa_{{l+1},i}(\x, \x') = \cov\left(h_{l+1, i}(\x), h_{l+1, i}(\x')\right)\notag \\
	& =  \iint w_l(\x, \z) \kappa_{l,i}(\z, \z')w_l(\z', \x')\d \z \d \z',  \label{eq:cov}\\
	&c_{l,i}(\x, \x') =\cov\left(h_{l+1, i}(\x), h_{l,i}(\x')\right) \notag \\
	&=  \int w_l(\x, \z) \kappa_{l,i}(\z, \x') \d \z. \label{eq:cross-cov}   
\end{align}

To perform  a \textit{nonlinear} transform, we instead model $h_{l+1,i}$ as the output of a nonlinear activation applied pointwise to $h_{l, i}$, where the activation function itself is drawn from a GP:  
\begin{align}
	h_{l+1, i}(\x) &= a_l\left(h_{l, i}(\x)\right), \;\;\; a_l \sim \gp\left(0, \vartheta_l(z, z')\right), \label{eq:gp-nonlinear-transform}
\end{align}
where $\vartheta_l(\cdot, \cdot)$ denotes the covariance function of the GP. Conditioning on $h_{l,i}(\cdot)$, the resulting process $h_{l+1,i}(\cdot)$ remains a GP, with covariance function given by the kernel composition: 
\begin{align}
	\kappa_{{l+1},i}(\x, \x')= \vartheta_{{l}}(h_{l,i}(\x), h_{l,i}(\x')). \label{eq:cov-nest}
\end{align}

%overall model, and joint probability distribution
%explain probabilistic framework, inducing point, and go on -- give audience a sense of probabilistic framework
\subsection{Model Framework}\label{sect:model-arch}
%\vspace{-0.1in}
In general, we assume the input function $f: \Omega \rightarrow \mathbb{R}^{d_0}$ is observed at a set of locations $\X_{\text{in}} = \{\x_{\text{in},j}\}_{j=1}^{N_{\text{in}}}$ and the output function $u: \Omega \rightarrow \mathbb{R}^{d_1}$ observed at $\X_{\text{out}} = \{\x_{\text{out},j}\}_{j=1}^{N_{\text{out}}}$. Notably,  $\X_{\text{in}}$ and $\X_{\text{out}}$ can be \textit{different} or even \textit{non-overlapping}. These sampling locations can be sparse and irregular, and may vary across different input-output function pairs during both training and inference.  

To flexibly accommodate varying sampling locations while enabling tractable inference thoughout  GP layers, we introduce a set of fixed locations $\X_Q$ 
 to serve as  \textit{projection points}. All latent functions are tracked and inferred consistently at these locations.
   We begin by placing independent GP priors on each component of the input function: $f^j \sim \mathcal{GP}(0, \nu_j(\cdot, \cdot))$ where  $f^j$ denotes the $j$-th component of $f$ with covariance function $\nu_j$\footnote{It is straightforward to extend this formulation to a multi-output GP prior over all components of $f$. We adopt independent priors here for simplicity and clarity of exposition.}.
   Let $f^j_Q$ denote the values of $f_j$ evaluated at $\X_Q$,  and let $\widehat{\f}^j$ denote its noisy observations  at $\X_{\text{in}}$. We define $\widehat{\F} = [\widehat{\f}^1,\ldots,\widehat{\f}^{d_0}]$ and $\F_Q = [\f^1_Q,\ldots,\f^{d_0}_Q]$. Their joint distribution factorizes as 
	\begin{align}
		&p(\widehat{\F}, \F_Q) = p(\widehat{\F})p(\F_Q|\widehat{\F}) \label{eq:prior-over-f} \\
		&= \prod\nolimits_{j} \N\left(\widehat{\f}_j | \0, \nu_j(\X_{\text{in}}, \X_{\text{in}})+ \sigma^2_j\I\right) \N(\f^j_Q|\m_j, \S_j), \notag
	\end{align}
where $\m_j = \nu_j(\X_Q, \X_{\text{in}})\K_j^{-1} \widehat{\f}_j$, $\S_j = \nu_j(\X_Q, \X_Q) - \nu_j(\X_Q, \X_{\text{in}})\K_j^{-1} \nu_j( \X_{\text{in}}, \X_Q)$, and $\K_j = \nu_j(\X_{\text{in}}, \X_{\text{in}}) + \sigma^2_j$,  with $\sigma^2_j$ denoting the observation noise variance. Detailed derivations are provided in Appendix Section~\ref{section:model-details}.

To further fuse information across input channels and enrich the latent representation, we introduce a learnable weight matrix $\W_0 \in \mathbb{R}^{d_0 \times C}$ to mix the $d_0$ components of the input function:  
$\H_1 = \F_Q \W_0 \in \mathbb{R}^{Q \times C}$. Each of the $C$ columns of $\H_1$ corresponds to a projected latent GP formed as a linear combination of the input components, evaluated at the shared projection locations $\X_Q$.

Next, we apply a sequence of linear and nonlinear transformations as described in Section~\ref{sect:gp-transform}. To simplify training and avoid the costly and complex computation of full conditional covariance matrices, we approximate the \textit{linear} transformation using the GP conditional mean (interpolation) rather than the full conditional Gaussian distribution. Specifically, we design a factorized conditional prior distribution: $p(\H_{l+1} \mid \H_l) = \prod_{i=1}^{C} p(\h_{l+1,i} \mid \h_{l,i})$, where
\begin{align}
	&p(\h_{l+1,i} \mid \h_{l,i}) \notag \\
	&= \delta\left(\h_{l+1,i} - c_{l,i}(\X_Q, \X_Q) \kappa_{l,i}(\X_Q, \X_Q)^{-1} \h_{l,i}\right).
	\label{eq:actual-linear-transform}
\end{align}
Here, $\delta(\cdot)$ denotes the Dirac delta distribution, and $\h_{l,i}$ and $\h_{l+1,i}$ denote the $i$-th columns of $\H_l$  and $\H_{l+1}$, respectively, corresponding to the projections of the latent functions $h_{l,i}(\cdot)$ and  $h_{l+1,i}(\cdot)$ at the  locations $\X_Q$.  This prior deterministically encodes the GP conditional mean of the linear transformation,  serving as a computationally efficient approximation, while retaining the expressiveness of GP-based functional transformations.  

To perform the \textit{nonlinear} transformation, for each GP activation $a_l(\cdot)$ defined in~\eqref{eq:gp-nonlinear-transform}, we introduce a set of inducing locations
$\bbeta = [\beta_1, \ldots, \beta_S]^\top \in \mathbb{R}$, with corresponding inducing values
$\boldeta_l = [a_l(\beta_1), \ldots, a_l(\beta_S)]^\top$. These inducing variables follow the GP prior $p(\boldeta_l) = \mathcal{N}\left(\boldeta_l \mid \0, \vartheta_l(\bbeta, \bbeta)\right)$.
Conditioned on  $\boldeta_l$, we define a factorized conditional prior:
\begin{align}
	&p(\H_{l+1} \mid \H_l, \boldeta_l) = \prod\nolimits_{\gamma_{l+1} \in \H_{l+1}} p(\gamma_{l+1} \mid \H_l, \boldeta_l) \notag \\
	&= \prod\nolimits_{\gamma_{l+1}\in \H_{l+1}} \mathcal{N}\left(\gamma_{l+1} \mid \mu_{lk}, t_{lk} \right),
	\label{eq:nonlinear-trans}
\end{align}
where $\gamma_{l+1}$ denotes an individual  element of $\H_{l+1}$ and $\gamma_l$ is the corresponding element in $\H_l$. The conditional mean and variance are given by  $\mu_{lk}= \vartheta_l\left(\gamma_l, \bbeta\right) \vartheta_l(\bbeta, \bbeta)^{-1} \boldeta_l$ and $t_{lk}= \vartheta_l\left(\gamma_l, \gamma_l\right) - \vartheta_l\left(\gamma_l, \bbeta\right)\vartheta_l(\bbeta, \bbeta)^{-1}\vartheta_l\left(\bbeta, \gamma_l\right)$. Together, the prior $p(\boldeta_l)$ and the conditional distribution in~\eqref{eq:nonlinear-trans} recover the classical sparse pseudo-input GP formulation~\citep{snelson2005sparse}. This construction avoids the expensive computation of the full covariance matrix (of size $C N_Q$) over $\H_{l+1}$, while retaining the ability to capture strong dependencies among its elements through the shared inducing variables.
%Note that the prior of $p(\boldeta)$ along with the conditional distribution~\eqref{eq:nonlinear-trans} form the classical sparse pseudo-input GP prior by~\citep{snelson2005sparse}. This design avoids the expensive computation of the full covariance matrice (of size $CN_Q$) between the elements of $\H_{l+1}$ while also preserving the (storng) correlations between these elements. 
%\begin{align}
%	p(\H_{l+1} \mid \H_l, \boldeta_l)
%	= \mathcal{N}\left(\vec(\H_{l+1}) \mid \bmu_l, \T_l \right),
%	\label{eq:nonlinear-trans}
%\end{align}
%where $\bmu_l= \vartheta_l\left(\vec(\H_l), \bbeta\right) \vartheta_l(\bbeta, \bbeta)^{-1} \boldeta_l$, $\vec(\cdot)$ denotes vectorization, and $\T_l= \vartheta_l\left(\vec(\H_l), \vec(\H_l)\right) - \vartheta_l\left(\vec(\H_l), \bbeta\right)\vartheta_l(\bbeta, \bbeta)^{-1}\vartheta_l\left(\bbeta, \vec(\H_l)\right)$.

The final layer $\H_L$ is obtained via a linear transform as in~\eqref{eq:actual-linear-transform}, but evaluated at the output observation locations $\X_{\text{out}}$. We then apply a learnable weight matrix $\W_1 \in \mathbb{R}^{C \times d_{1}}$ to aggregate the $C$ latent channels into the output space: $\U = \H_L \W_1 \in \mathbb{R}^{N_{\text{out}} \times d_1}$. Let $\Y$ denote the observed outputs. We adopt a Gaussian likelihood, $p(\Y|\U) = \prod_{i=1}^{d_1} \N(\y_i|\u_i, v_i\I)$, where $\y_i$ is the $i$-th component of the output function observed at the $N_{\text{out}}$ locations, $\u_i$ the $i$-th column of $\U$, and $v_i$ the corresponding noise variance. Appendix Section~\ref{sect:cond-gp-priors} provides further clarification of the conditional GP priors induced by our construction in function space.

%We augment the model with the inducing variables \( \boldeta_l \) at each GP layer \( l \) that performs a nonlinear transformation. Using GP interpolation, we compute the values of the transformed function \( h_{l+1,i}(\cdot) \) at the projection points \( \X_{Q} \). According to~\eqref{eq:gp-int}, we obtain
%%\[
%$\h_{l+1,i} = \vartheta_{l}(\h_{l, i}, \bbeta)\, \vartheta_{l}(\bbeta, \bbeta)^{-1} \boldeta_l$,
%%\]
%where \( \h_{l,i} \) denotes the values of \( h_{l,i}(\cdot) \) at the projection points \( \X_Q \).

%A sequence of linear and nonlinear transformations is then applied, as defined in~\eqref{eq:gp-linear-transform} and~\eqref{eq:gp-nonlinear-transform}.
% At the last layer $\H_L$, we apply a weight matrix $\W_1 \in \mathbb{R}^{C \times d_{1}}$ to compute the output function prediction at $\X_Q$: $\widehat{\U} =  \widehat{\H}_{L}  \W_1 \in \mathbb{R}^{Q \times d_1}$. We then use GP to interpolate the prediction to $\X_{\text{out}}$,
% \begin{align}
% 	\U = [\u_{1}, \ldots, \u_{d_1}], \;\;\;\; \u_j =\rho_j(\X_{\text{out}}, \X_Q)\rho_j(\X_Q, \X_Q)^{-1}  \widehat{\u}_j, \label{eq:int-output}
% \end{align} 
% where $\widehat{\u}_j$ is the $j$-th column of $\widehat{\U}$, representing the prediction of the $j$-th component of $u$ at $\X_{Q}$,  and $\rho_j(\cdot, \cdot)$ is the associated kernel function. The model architecture is visualized in Appendix Figure~\ref{fig:model-arch}.

\section{Algorithm}
\subsection{Discrete Approximation of Integral Transforms}

A critical challenge in implementing our model lies in computing and tracking the covariance function $\kappa_{l,i}(\cdot, \cdot)$ for each GP layer ($l = 1, 2, \ldots$) and the cross-covariance function $c_{l,i}(\cdot, \cdot)$ required during linear transformations. These computations involve repeated integral transforms (see~\eqref{eq:cov} and~\eqref{eq:cross-cov}) and nested compositions (see~\eqref{eq:cov-nest}), making closed-form derivations intractable and resulting procedures computationally demanding. To address this challenge, we adopt a discrete approximation of the integral transform for computing the cross-covariance (see~\eqref{eq:cross-cov}). This leads to a striking insight: the covariance and cross-covariance matrices \textit{cancel out}, dramatically simplifying the implementation.

Specifically, we consider a quadrature rule to approximate the integral transform. Let $\balpha = [\alpha_1, \ldots, \alpha_M]^\top$ denote the quadrature weights and $\ol{\X} = \{\ol{\x}_j\}_{j=1}^M$ the corresponding nodes. The cross-covariance function in~\eqref{eq:cross-cov} is then approximated as
\begin{align}
	c_{l,i}(\x, \x') \approx \sum\nolimits_{m=1}^M \alpha_m \cdot w_l(\x, \ol{\x}_m)\kappa_{l,i}(\ol{\x}_m, \x'). \label{eq:quad-rule}
\end{align}
According to~\eqref{eq:actual-linear-transform}, the values of $h_{l+1,i}(\cdot)$ at the projection points \(\X_Q\) are given by
\begin{align}
	\h_{l+1,i} = c_{l,i}(\X_{Q}, \X_{Q})\kappa_{l,i}(\X_{Q}, \X_{Q})^{-1} \h_{l, i}. \label{eq:quad-int-cross-cov}
\end{align}
Substituting~\eqref{eq:quad-rule} into~\eqref{eq:quad-int-cross-cov}, we obtain: $\h_{l+1,i} = \W_l \cdot \diag(\balpha)  \cdot \kappa_{l,i}(\ol{\X}, \X_{Q}) \kappa_{l,i}(\X_{Q}, \X_{Q})^{-1} \h_{l, i}$, 
%\begin{align}
%	\h_{l+1,i} = \W_l \cdot \diag(\balpha)  \cdot \kappa_{l,i}(\ol{\X}, \X_{Q}) \kappa_{l,i}(\X_{Q}, \X_{Q})^{-1} \h_{l, i},
%\end{align}
where $\W_l$ denotes the evaluations of the weight function $w_l(\cdot, \cdot)$ over the Cartesian product $\X_Q \times \ol{\X}$. Now, if we set the projection points $\X_Q = \ol{\X}$, the cross-covariance and covariance matrices cancel out, yielding:
\begin{align}
	\h_{l+1,i} = \W_l \cdot \diag(\balpha) \cdot \h_{l, i}, \label{eq:linear-transform}
\end{align}
which is a discrete approximation applied directly to the integral transform of \(h_{l,i}(\cdot)\) as defined in~\eqref{eq:int-transform}. 

This observation holds for any discrete approximation for the cross-covariance function~\eqref{eq:cross-cov}: as long as we set the projection points $\X_{Q}$ to the locations used for the approximation, the GP interpolation in ~\eqref{eq:actual-linear-transform} reduces to the form of ~\eqref{eq:linear-transform}. That means, we never need to explicitly compute or track the covariance function at intermediate GP layers. The only required covariance functions are: $\nu_j(\cdot, \cdot)$ for the input function $f$ (see~\eqref{eq:prior-over-f}) and $\vartheta_{l}$ for constructing each GP activation (see~\eqref{eq:gp-nonlinear-transform}). This insight not only streamlines the model implementation, but also enables flexible choices of integral transform approximation.

We consider two approximation methods. The first follows the quadrature-based approach described in~\eqref{eq:quad-rule} and~\eqref{eq:linear-transform}, in which we learn $\W_l$---the discretized values of $w_l(\cdot)$ evaluted on the grid \(\X_Q \times \X_Q\). However, because the input to $w_l$ has twice the dimensionality of the latent function
 $h_{l,i}(\x)$, the resulting matrix  $\W_l$ can be large, leading to high computational and memory costs. To address this issue, we introduce a \textit{dimension-wise integral transform}. For illustration, consider a two-dimensional input $\x = [x_1, x_2] \in \mathbb{R}^2$. We define two separate weight functions, \(w_l^1(x_1, x_1')\) and \(w_l^2(x_2, x_2')\), and design the transformation as:
\begin{align}
	&h_{l+1, i}(x_1,  x_2) = \int w_l^1(x_1, x_1') h_{l, i}(x_1', x_2) \d x_1' \notag \\
	&+ \int w_l^2(x_2, x_2') h_{l, i}(x_1, x_2' ) \d x_2'. \label{eq:dimwise}
\end{align} 
We approximate each integral in~\eqref{eq:dimwise} using 1D quadrature rules. As a result, we only need to estimate two 2D matrices, \(\W_l^1\) and \(\W_l^2\), which represent the values of $w_l^1(\cdot, \cdot)$ and $w_l^2(\cdot, \cdot)$ over the Cartesian products of the respective quadrature nodes. 

For the second approximation, we leverage ideas from Fourier Neural Operators~\citep{li2020fourier,tran2021factorized}. We assume each weight function is stationary, i.e., $w_l^j(x_j, x'_j) = w_l^j(x_j - x'_j)$, and place the projection points $\X_Q$ on a regular grid in each dimension. Using the convolution theorem~\citep{bracewell1966fourier}, each integration in R.H.S of ~\eqref{eq:dimwise} can be expressed as: 
\begin{align}
	 &\int w_l^1(x_1, x_1') h_{l, i}(x_1', x_2) \d x_1' = \int w_l^1(x_1 - x_1') h_{l, i}(x_1', x_2) \d x_1' \notag \\
	 &=  \mathcal{F}^{-1}\left[\mathcal{F}[w_l^1(\cdot)] \cdot \mathcal{F}[h_{l,i}(\cdot, x_2)]\right], \label{eq:conv} \\
	 &\int w_l^2(x_2, x_2') h_{l, i}(x_1, x_2' ) \d x_2' =  \mathcal{F}^{-1}\left[\mathcal{F}[w_l^2(\cdot)] \cdot \mathcal{F}[h_{l,i}(x_1, \cdot)]\right], \notag 
	% \int w_l^j(x_j, x_j') h_{l, i}(x_1', x_2) \d x_1' = \int w_l(\x - \x') h_{l,i}(\x') \, \mathrm{d}\x' = \mathcal{F}^{-1}\left[\mathcal{F}[w_l] \cdot \mathcal{F}[h_{l,i}]\right], \label{eq:conv}
\end{align}
where \(\mathcal{F}[\cdot]\) and \(\mathcal{F}^{-1}[\cdot]\) denote the Fourier and inverse Fourier transforms, respectively. In practice, we approximate~\eqref{eq:conv} by applying the discrete Fourier transform (DFT) to $\h_{l,i}$ evaluated on $\X_Q$ along each dimension, multiplying the result with the discretized spectrum of $w_l^1, w_l^2, \ldots$, and then applying the inverse DFT to obtain \(\h_{l+1,i}\).

While this dimension-wise approximate integral transform may bring up additional errors relative to~\eqref{eq:int-transform}, it substantially reduces the number of mode parameters and alleviates the risk of overfitting in practice.  Appendix Section~\ref{sect:error-analysis} provides a mathematical analysis of the resulting discretization error, and ablation studies in Appendix Section~\ref{sect:ablation} further validate the empirical benefits of this approach.

\subsection{Stochastic Variational Inference}
To enable probabilistic inference, we develop an efficient stochastic variational inference (SVI) algorithm~\citep{wainwright2008graphical,hensman2013gaussian}. Let $\widehat{\F}$ and $\Y$ denote an observed pair of input and output functions. As described in Section~\ref{sect:model-arch}, the joint probability of our model can be written as
\begin{align}
	&p(\text{joint}) = p(\widehat{\F})p(\F_Q|\widehat{\F}) \cdot \prod\nolimits_{l+1 \in \Gamma_{\text{lin}}} p(\H_{l+1}|\H_l) p(\Wcal_l) \notag \\
	&\cdot \prod\nolimits_{l+1 \in \Gamma_{\text{non}}} p(\boldeta_l) p(\H_{l+1}|\H_l, \boldeta_l) \cdot p(\Y|\U),  \notag 
\end{align}
where $\Gamma_{\text{lin}}$ and $\Gamma_{\text{non}}$ denote the set of linear and nonlinear layers, respectively, and $p(\Wcal_l)$ is the prior over the weight function values.  For clarity of exposition, we present the method using a single training instance, though the extension to multiple instances is straightforward. Given the training dataset $\Dcal$, our goal is to infer the posterior distribution of the inducing variables $\{\boldeta_l\}$ together with the point estimates of the kernel parameters and weight function values $\Wcal_l$, and noise variances. 
At prediction time, we sample $\{\boldeta_l\}$ from their posterior distribution and propagate these samples through the model to generate samples of $\H_l$ across layers, ultimately producing predictive samples of $\Ucal$ and enabling principled uncertainty quantification.

Direct inference over ${\boldeta_l}$ is intractable since the posterior distribution has no closed-form solution.  To address this issue, we use the variational inference framework~\citep{wainwright2008graphical}.  We first apply a whitening transformation~\citep{murray2010slice}:
\begin{align}
	\boldeta_l = \A_l \ol{\boldeta}_l, \quad p(\ol{\boldeta}_l) = \mathcal{N}(\0, \I), \quad \A_l \A_l^\top = \vartheta_{l}(\bbeta, \bbeta), \notag
\end{align} 
where \( \A_l \) is the Cholesky factor of the covariance matrix $\vartheta_{l}(\bbeta, \bbeta)$ defined over the inducing points, whose dimensionality is small. The standard Gaussian prior over $\ol{\boldeta}_l$ follows directly from the original prior  $p(\boldeta_l) = \N(\0, \vartheta_{l}(\bbeta, \bbeta))$. This reparameterization decouples the latent variables from the kernel parameters, thereby simplifying posterior inference. 
We then construct a variational approximation: 
\begin{align}
	&p(\{\ol{\boldeta}_l, \H_{l+1}\}|\Dcal) \approx q(\{\ol{\boldeta}_l, \H_{l+1}\}) \notag \\
	&\propto \prod\nolimits_{l+1 \in \Scal_{\text{non}}} q(\ol{\boldeta}_l) p(\H_{l+1}|\H_l, \A_l \ol{\boldeta}_l). \notag
\end{align}
Following the variational inference framework, the evidence lower bound (ELBO),  $\Lcal = \EE_{q}\left[\frac{p(\text{Joint})}{q(\{\ol{\boldeta}_l\})}\right]$, can be written as  
\begin{align}
	&\Lcal = \EE_q\left[\p(\widehat{\F})p(\F_Q|\widehat{\F}) \prod\nolimits_{l+1 \in \Scal_{\text{lin}}} p(\H_{l+1}|\H_l) p(\Wcal_l)\right] +\notag \\
	& \EE_q\left[\frac{\prod_{l+1 \in \Scal_{\text{non}}} p(\ol{\boldeta}_l) \bcancel{p(\H_{l+1}|\H_l, \A_l \ol{\boldeta}_l)} \cdot p(\Y|\U)}{\prod\nolimits_{l+1 \in \Scal_{\text{non}}} q(\ol{\boldeta}_l) \bcancel{p(\H_{l+1}|\H_l, \A_l \ol{\boldeta}_l)}}\right]. \notag 
\end{align}
Notably, all conditional terms $p(\H_{l+1}|\H_l, \A_l \ol{\boldeta_l})$ cancel analytically, yielding a tractable and simplified ELBO.  We use the reparameterization trick~\citep{kingma2013auto} to compute unbiased Monte Carlo estimates of the expectations and their gradient, and optimize the ELBO using stochastic gradient descent.

\noindent\textbf{Computational Complexity}: The time complexity of our training method is $\Ocal(dN_Q^2BL + S^3)$ for the quadrature-based dimension-wise integral transform~\eqref{eq:dimwise} and $\Ocal(dN_Q\log N_QBL + S^3)$ for the Fourier transform approach~\eqref{eq:conv} (via FFT). Here, $d$ is the input dimension, $N_Q$ the number of quadrature nodes or sampling collocations, $L$ the number of layers, $S$ the number of inducing points, and $B$ is the mini-batch size for stochastic training.   In both cases,  the time complexity scales linearly with input dimensionality. The space complexity  is $O(L(dN_Q^2 + BN_QC + S^2))$ for the quadrature-based approach, and $O(L(dN_Q + BN_QC + S^2))$ for the Fourier-based method, accounting for the storage of weight function values, the hidden function values $\Hcal_i$, and the variational posterior $q(\{\boldeta_l\})$.

%LFR
\section{Related Work}

Functional data analysis (FDA) has been a prominent area of statistical research for several decades, tracing back to foundational works such as \citet{ramsay1991some,faraway1997regression}. Key topics in this field include functional regression~\citep{morris2015functional} and functional principal component analysis (FPCA)~\citep{silverman1996smoothed,hall2006properties}.  A major subfield is function-on-function regression, where both the predictors and the response are functional. Other widely studied formulations include scalar-on-function regression~\citep{goldsmith2014estimator,reiss2017methods,hullait2021robust}, function-on-scalar regression~\citep{reiss2010fast,bauer2018introduction}, as well as hybrid or mixed cases. A variety of function-on-function regression methods have been developed~\citep{yao2005functional,manrique2016functional,kim2018additive,luo2019interaction,beyaztas2020function,aneiros2022variable,wang2022functional,dette2024pivotal}, with most of them grounded in the functional linear regression framework, an intuitive extension of classical linear regression. These methods primarily differ in their choice of basis functions, normalization, regularization, \etc 
%used to represent the predictor, response, and the weight functions, normalization, regularization, and smoothing techniques.% Additional distinctions arise in their use of normalization, regularization, and smoothing techniques.

In scientific machine learning, operator learning has emerged as an vibrant  field aimed at learning mappings between function spaces governed by PDEs. These mappings --- often referred to as operators --- represent relationships involving derivatives and integrals. Many neural architectures have been specifically designed for learning such PDE operators. One most prominent class is the Fourier Neural Operators (FNO)~\citep{li2020fourier} and their extensions~\citep{tran2021factorized, lingsch2024beyond}, which perform functional transformations through Fourier layers combined with standard neural network activations such as GeLU. Other notable  models include the Multiwavelet Neural Operator~\citep{gupta2021multiwavelet}, CNO~\citep{raonic2024convolutional}, DeepONet~\citep{lu2021learning}, and transformer-based approaches~\citep{cao2021choose, li2022transformer, hao2023gnot}, among others.
%One class of most successful examples are the Fourier Neural Operators (FNO)~\citep{li2020fourier} and its variants, such as~\citep{tran2021factorized, lingsch2024beyond} , which performs functional transformations using Fourier layers combined with standard neural network activations such as GeLU. Other NO models include Multiwavelet Neural Operator~\citep{gupta2021multiwavelet},   CNO~\citep{raonic2024convolutional},  DeepONet~\citep{lu2021learning}, transformer-based NO's~\citep{cao2021choose,li2022transformer,hao2023gnot}, and others. 
%More recently, \citet{li2024multi} introduced active learning methods tailored for multi-resolution FNOs. %Another  approach is the Deep Operator Network %(DeepONet)~\citep{lu2021learning}, which comprises a branch network (operating on the input function values) and a trunk network (applied to the sampling locations), with predictions formed via the dot product of the two outputs. To enhance stability and computational efficiency, \citet{lu2022comprehensive} proposed replacing the trunk network with Proper Orthogonal Decomposition (POD, i.e., PCA) bases. 
%Transformer-based models have also been explored for operator learning~\citep{cao2021choose,li2022transformer,hao2023gnot}, leveraging attention mechanisms to capture complex dependencies. %Parallel to these neural architectures, recent work has also investigated kernel-based operator learning approaches~\citep{long2022kernel,batlle2023kernel}, offering alternative probabilistic and nonparametric formulations.

The classical Deep GP framework~\citep{damianou2013deep}, is designed to learn a single function from its observed values across various input locations, aligning with standard GP regression setting. Our formulation extends this paradigm by considering both the input and output as functions, aiming to learn their relationship directly in the functional space. 
Our variational inference is similar to~\citep{salimbeni2017doubly}, where for each GP activation, we introduce  a set of inducing variables  to facilitate tractable latent function estimation. However, we  incorporate a whitening transformation that decouples the inducing variables from the kernel hyperparameters to further faciliate the learning and inference.  

%The classical Deep Gaussian process framework, as formulated by~\citet{damianou2013deep}, is designed for learning a single function from its observed function values across different locations. The task is the same as standard GP regression.  Our formulation differs in that  both the input and output are functions, and our model aims to capture their relationship in the functional space. To this end, we model the input function, output function, as well as a sequence of linear transformation and nonlinear activations as GPs. Our sparse variational inference is similar to~\citep{salimbeni2017doubly}, where for each GP activation, we  introduce a set of inducing locations and values to create function estimation.  However, we further leverage the whitening trick~\citep{murray2010slice} to convert the prior over the transformed inducing values into a standard Gaussian distribution. In this way, we can decouple the strong correlation between the kernel parameters and inducing values, making the optimization of the variational free energy easier.  

%\vspace{-0.1in}
\section{Experiment}
\vspace{-0.05in}
%We evaluated \ours on three numerical PDE simulation scenarios and three real-world applications. The PDE simulation scenarios represent the central focus  of neural operator methods. To further demonstrate the versatility of our approach, we extended our evaluation to real-world applications characterized by sparse, noisy, and irregularly sampled data. This comprehensive evaluation examines the robustness and adaptability of our method across both controlled and practical settings.
We evaluated \ours on three synthetic datasets and three real-world applications with sparse, noisy, and irregularly sampled data.

The simulation scenarios are as follows. (1) \textbf{Burgers}~\citep{lu2022comprehensive}:  learning a mapping from the initial condition $u_0(x)$ to the solution of a Burger's equation at time $t = 1$, denoted by $u_1(x)$, where $u_0, u_1 : (0, 1) \rightarrow \mathbb{R}$. (2) \textbf{Darcy Flow}~\citep{lu2022comprehensive},  predicting the pressure field $u: [0, 1]^2 \rightarrow \mathbb{R}$ from the given permeability field $c: [0, 1]^2 \rightarrow \mathbb{R}$, governed by the Darcy flow equation. (3) \textbf{Car Shape}~\citep{umetani2018learning}: learning the mapping from the three-dimensional  car geometry to the corresponding air-pressure field under a fixed inlet velocity. Both the input (shape) and output (pressure) functions are discretized using triangular meshes. 
For \textbf{Burgers} and \textbf{Darcy Flow}, we used 1,000 training examples and 200 test examples. For \textbf{Car Shape}, we employed 500 training examples, and 278 testing examples . Additional dataset details are provided in Appendix Section~\ref{sect:dataset}.

We evalauted \ours on the following real-world applications. (1)  \textbf{Beijing-Air}\footnote{\url{https://archive.ics.uci.edu/dataset/501/beijing+multi+site+air+quality+data}}, including hourly measurements of multiple air pollutants in Beijing from 2014 to 2017. The task is to predict the hourly concentration of CO over the subsequent week using the polutent measurements from the preceeding week.  We randomly selected 5,000 weekly sequences for training and 1,000 for testing.  (2) \textbf{SLC-Precipitation}, 
compiled from daily precipitation records collected by weather stations distributed across the Great Salt Lake region between 1954 and 2023. The goal is to infer the next day's precipitation conditioned on observations from the current day. The weather stations are sparsely and irregularly distributed, with substantial missing data. We randomly selected 128 stations and constructed the task as a spatial prediction problem over functions. A total of 5,000 examples were used for training and 1,000 for testing.
 (3) \textbf{Quasar Reverberation Mapping}, derived from observations collected by the Zwicky Transient Facility (ZTF) at the Palomar Observatory~\citep{2019PASP..131a8002B}. The objective is to model the relationship between a quasar's central continuum emission and the delayed response from surrounding emitting regions, a phenomenon known as reverberation mapping~\citep{1982ApJ...255..419B,1993PASP..105..247P}. The dataset consists of 793 pairs of irregularly sampled light curves with non-aligned observation times between inputs and outputs. We used 650 examples for training and 143 for testing. Additional details are provided in Appendix Section~\ref{sect:lights}.

We compared \ours with the following methods: (1) Functional Linear  Regression (FLR)~\citep{morris2015functional}, the mainstream  function-on-function regression method that extends linear regression into functional spaces.  We employed two popular basis expansions: one with Fourier bases (denoted as LFR-Fourier) and the other with B-splines (LFR-BSpline). (2) Fourier Neural Operator (FNO)~\citep{li2022fourier}, the most widely used neural operator model that introduces Fourier layers to perform linear functional transformations. However, FNO requires inputs and output functions to be sampled on a uniform grid due to its reliance on Fast Fourier Transform (FFT). %(3) Factorized FNO (F-FNO)~\citep{tran2021factorized}, an extension of FNO designed to handle irregularly sampled data and reduce memory consumption. F-FNO first maps irregular spatial locations to a regular latent grid via a neural network, then applies Fourier layers along each input dimension. 
(3) DSE-FNO~\citep{lingsch2024beyond}, a most recent variant of FNO, which uses non-uniform discrete Fourier transforms (NUDFT) to enable direct spectral evaluations on irregular domains. However, DSE-FNO  still assumes that both input and output functions share identical sampling locations to main the consistency between NUDFT and its inverse. (4) GNOT~\citep{hao2023gnot}, a transformer-based neural operator capable of flexibly handling arbitrarily irregular sampling in both input and output functions via cross-attention mechanisms~\citep{vaswani2017attention}. (5) LFR-GP, a baseline model that uses a single integral GP transform layer with Gauss-Legendre quadrature (see~\eqref{eq:linear-transform}), effectively representing a simplified version of \ours with one GP layer.
We evaluated two versions of our method: \ours-QR, which performs dimension-wise integral transforms using numerical quadrature rules (see~\eqref{eq:dimwise}); \ours-FT, which leverages the convolution theorem and Fourier transform to compute the integral transforms (see~\eqref{eq:conv}). For each method on each task, we used a separate validation set to select the optimal hyperparameters based on one training dataset. These hyperparameters were then fixed, and the model was trained and tested across five independent runs. We report both the mean prediction error and the standard deviation to assess performance stability and accuracy.  We leave the details about the hyperparameter selection for each method in Appendix Section~\ref{sect:hyper}.

%For \ours, we used the ADAM optimizer, with the learning rate selected from \{1e-5, 1e-4, 5e-4, 1e-3, 5e-3, 1e-2, 1e-1\}. The maximum number of training epochs was set to XXX. The number of GP layers was varied in \{2, 3, 4\}. The kernel functions for both input and output GPs, as well as the GP activation, were chosen from the Mat\'ern family, with degrees of freedom $1+p/2$ where $p\in \{0, 1, 2, 3, 4\}$, or their weighted combinations.  Each GP activation used XX inducing points. The number of latent channels was varied in XXX. For \ours-QR, the number of quadrature nodes was selected from XXX; for \ours-FT, the number of latent spatial locations was selected from XXX.  Hyperparameter tuning procedures for all other baselines are provided in Appendix Section~XX. For each method on each task, we used a separate validation set to select the optimal hyperparameters based on one training dataset. These hyperparameters were then fixed, and the model was trained and tested across five independent runs. We report both the mean prediction error and the standard deviation to assess performance stability and accuracy.

\begin{table*}[!htp]
	\centering
	\caption[my-caption]{\small Normalized Root-Mean-Square (NRMSE) error. The top two results are highlighted in bold. N/A  indicates that the method is not applicable\protect\footnotemark.}\label{tab:nrmse}
	\vspace{-0.05in}
	\small
	\begin{subtable}{\linewidth}
		\centering
		\small 
		\caption{\small Synthetic datasets.}\label{tab:nrmse-simu}
		\vspace{-0.05in}
	\begin{tabular}{cccc}\toprule
		\textit{Method} & \textit{Burgers} & \textit{Darcy} & \textit{Car Shape}  \\\midrule
		FLR-Fourier &3.76e-1 $\pm$ 1.90e-3 & 4.58e-1 $\pm$ 6e-4  & 9.86e-1  $\pm$ 1.0e-4  \\
		FLR-BSpline &4.08e-1 $\pm$ 2.00e-3 & 4.54e-1 $\pm$ 6e-4  & 9.92e-1 $\pm$ 1.0e-4  \\
		FLR-GP & 3.36e-1 $\pm$ 1.00e-3 & 5.54e-1 $\pm$ 1.70e-3 & 6.16e-1 $\pm$ 2.01e-2 \\
		GNOT &8.90e-3 $\pm$ 1.40e-4 & 2.58e-2 $\pm$ 3.40e-4 & \textbf{1.03e-1 $\pm$ 9.8e-4}  \\
		FNO & \textbf{2.76e-3 $\pm$ 4.17e-5} & \textbf{1.79e-2 $\pm$ 2.49e-4} & N/A  \\
	%	F-FNO & \textbf{2.76e-3 $\pm$ 4.17e-5} &\textbf{1.75e-2 $\pm$ 1.67e-4} & 5.76e-1 $\pm$ 4.73e-2  \\
		DSE-FNO & 6.69e-2 $\pm$ 9.89e-4 & 3.63e-2 $\pm$ 1.70e-4 &  3.56e-1 $\pm$ 3e-2    \\
		\ours-FT & \textbf{1.79e-3 $\pm$ 1.86e-4} & \textbf{1.67e-2 $\pm$ 1.40e-4} &  1.22e-1 $\pm$ 4.15e-2 \\
		\ours-QR &5.61e-3 $\pm$ 6.58 e-4&1.83e-2 $\pm$ 5.17e-4 &  \textbf{1.15e-1 $\pm$ 1.11e-2}  \\
		\bottomrule
	\end{tabular}
	\end{subtable}
	\begin{subtable}{\linewidth}
		\centering
		\caption{\small Real-world datasets.}\label{tab:nrmse-real}
		\vspace{-0.05in}
		\begin{tabular}{cccc}\toprule
			\textit{Method} 	&\textit{Beijing-Air} & \textit{Quasar} & \textit{SLC-Precipitation} \\\midrule
			FLR-Fourier &0.636 $\pm$ 3.00e-3 & 0.008 $\pm$ 1.00e-3 & 1.02 $\pm$ 3.40e-3 \\
			FLR-BSpline & 0.639 $\pm$ 3.00e-3 & 0.008 $\pm$ 1.00e-4 &1.02 $\pm$ 3.50e-3 \\
			FLR-GP & 0.552 $\pm$ 1.90e-3 & 0.0079 $\pm$ 1.00e-4 & 1.03 $\pm$ 1.06e-3\\
			GNOT & 0.553 $\pm$ 1.80e-3 & 0.0054 $\pm$ 1.00e-4 & 0.836 $\pm$ 6.76e-3 \\
			FNO & 0.403 $\pm$ 2.10e-3 & N/A  &N/A  \\
	%		F-FNO & 0.403 $\pm$ 2.10e-3 & 0.441 $\pm$ 8.20e-3 & 0.903 $\pm$ 5.15e-3 \\
			DSE-FNO & 0.529 $\pm$ 1.34e-3 &N/A  &N/A \\
			\ours-FT & \textbf{0.304 $\pm$ 1.16e-3} & \textbf{0.0047 $\pm$ 6.06e-5} &  \textbf{0.776 $\pm$ 4.64e-3} \\
			\ours-QR & \textbf{0.201 $\pm$ 5.64e-3} & \textbf{0.0044 $\pm$ 5.80e-5} &  \textbf{0.773 $\pm$ 4.52e-3} \\		
			\bottomrule
		\end{tabular}
	\end{subtable}
\end{table*}
\footnotetext{FNO is not applicable to irregularly sampled locations, while DSE-FNO cannot be used when the input and output functions are sampled at different locations, leading to missing results on several datasets. \cmt{Since the official DSE-FNO implementation does not support 3D functional mappings (required in \textit{3D-NS} ), we developed our own version. However, it produced unreasonably large errors, and thus the corresponding results are omitted.} }

\begin{table*}[t]\centering
	\caption{\small Mean Negative Log-Likelihood (MNLL)\protect\footnotemark. The top two results are highlighted in bold.}\label{tab:nll}
	\vspace{-0.05in}
	\small
	\begin{subtable}{\linewidth}
		\centering
		\small 
			\caption{\small Synthetic datasets.}\label{tab:nll-simu}
			\vspace{-0.05in}
		\begin{tabular}{cccc}\toprule
			\textit{Method} & \textit{Burgers} & \textit{Darcy} & \textit{Car Shape}  \\ \midrule
			GNOT-MCDropout &2.02 $\pm$0.031 &2.84 $\pm$0.104 & 8.31 $\pm$ 0.241  \\
			FNO-MCDropout &312.46 $\pm$ 19.32 & 111.11  $\pm$4.10 & 24.34 $\pm$ 5.14  \\
			GNOT-SGLD & 5.33 $\pm$3.50e-3 & 13.29 $\pm$ 0.022 & 10.83 $\pm$ 0.135  \\
			FNO-SGLD &12.05 $\pm$2.96 & 13.24 $\pm$ 4.62 & {7.42 $\pm$ 1.58}  \\
			\ours-FT &\textbf{-4.27 $\pm$0.101} &\textbf{-3.50 $\pm$ 0.025} & \textbf{4.76 $\pm$ 0.813}  \\
			\ours-QR &\textbf{-3.29 $\pm$ 0.066} & \textbf{-4.13 $\pm$ 0.042}  & \textbf{1.37 $\pm$ 0.721}  \\
			\bottomrule
		\end{tabular}
\end{subtable}
	\begin{subtable}{\linewidth}
	\centering
	\small 
	\caption{\small Real-world datasets.}\label{tab:nll-real}
	\vspace{-0.05in}
		\begin{tabular}{cccc}\toprule
		\textit{Method} & \textit{Beijing-Air} & \textit{Quasar}  & \textit{SLC-Precipitation} \\ \midrule
		GNOT-MCDropout & 464.7$\pm$ 35.1 & 54.3 $\pm$5.97 & 404.5 $\pm$ 16.9 \\
		FNO-MCDropout & 171.1 $\pm $ 6.12 & N/A  &N/A  \\
		GNOT-SGLD & 10.38  $\pm$ 1.10e-3 & 17.06 $\pm$ 0.03 & 5298.74 $\pm$ 943.705 \\
		FNO-SGLD & 18.66 $\pm$ 3.08 & N/A  &N/A  \\
		\ours-FT & \textbf{7.78 $\pm$ 5.81e-2} & \textbf{-0.534 $\pm$6.69e-2} & \textbf{17.3 $\pm$ 2.18} \\
		\ours-QR & \textbf{8.01 $\pm$ 9.36e-2} & \textbf{-0.924 $\pm$1.71e-2} & \textbf{25.9 $\pm$ 4.82} \\
		\bottomrule
	\end{tabular}
\end{subtable}
\end{table*}
\footnotetext{The standard FNO is not applicable to the Car-Shape dataset. We therefore evaluated MNLL using the DSE-FNO variant.}

%\subsection{Predictive Performance}
\noindent\textbf{Prediction Accuracy}. We first evaluated the normalized root-mean-square error (NRMSE) for each method. 
As shown in Table~\ref{tab:nrmse}, our method (\ours-FT/-QR) consistently achieves the highest prediction accuracy, except on the Car-Shape dataset, where \ours-QR ranks second only to GNOT. 
On the \textit{Burgers} and \textit{Darcy} datasets, \ours-FT surpasses highly optimized neural operator models, while \ours-QR achieves comparable error levels. Notably, across all three real-world applications, both \ours-FT and \ours-QR significantly outperform all competing methods, with statistical significance exceeding the 95\% confidence level. Furthermore, neural operators such as FNO and DSE-FNO rely on regular or fixed sampling locations, which limits their applicability to  real-world scenarios with arbitrary observation points. For example, neither FNO nor DSE-FNO can be applied to the Quasar and SLC-Precipitation datasets, where the number and locations of sampling points vary not only between the input and output functions but also across different training and testing instances. Together the results demonstrate the strong flexibility and predictive performance of \ours.

%We first examined the normalized root-mean-square error (NRMSE) of each method. As shown in Table~\ref{tab:nrmse}, our method  (\ours-FT and/or \ours-QR) consistently achieves the top prediction accuracy, except in the NS dataset, FLR-Fourier and FLR-BSpline outperforms all the methods. This might be because the NS dataset is challenging (with a high March number) yet the number of training examples is quite limited (only 90 training examples are available). Under such cases, A linear modeling structure with simple basis functions can achieve more robust predictive performance. However, \ours still outperforms all the other methods by a large margin. In Burgers and Darcy datasets, \ours-FT can outperform highly optimized neural operator models, which \ours-QR achieves errors at the same level. it is worth-noting that on all the three real-world applications, \ours (both \ours-FT and \ours-QR) outperform all the competing methods substantially, with a significance level over 95\%. In addition, neural operators like FNO and DSE-FNO have the restrictions on the domain and sampling locations, and cannot apply to our real-world scenarios including arbitrarily observed locations. For example, Both FNO and DSE-FNO cannot be applied to the Quasar and SLC-Precipitation datasets because the sampling locations and their numbers vary not only between the pair of the input and output functions but also among different training and testing instances. Together these highlight the flexibility of \ours in real-world applications, and its outstanding predictive performance. 
\begin{figure*}%[h]
	\centering
	\includegraphics[width=\linewidth]{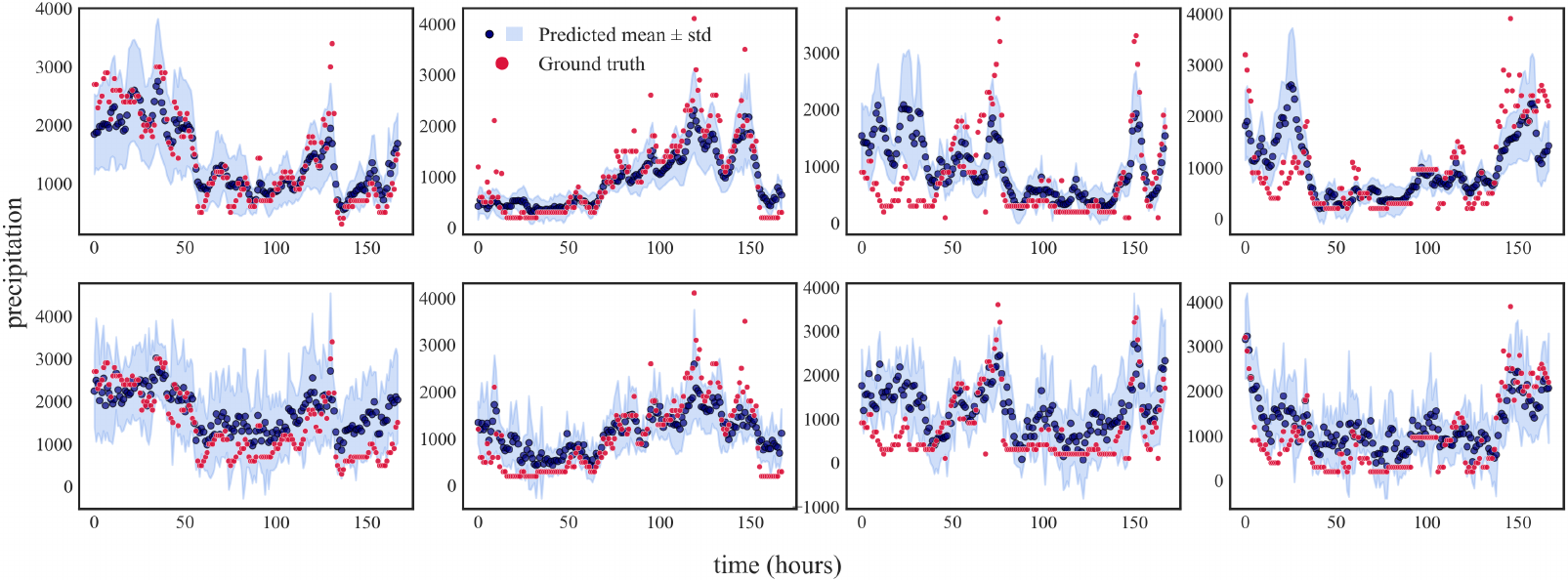}
	\caption{\small Prediction examples of \ours on \textit{Beijing-Air} dataset. The shaded regions indicate one predictive standard deviation. The top row shows the prediction of \ours-FT and the bottom row \ours-QR. } 
	\label{fig:model-pred-beijing}
	\vspace{-0.15in}
\end{figure*}
%\begin{figure}
%	\centering
%	\includegraphics[angle=270, width=1.\linewidth]{figs/darcy_examples_pair_3-trim.pdf}
%	\caption{\small Prediction examples of \ours-QR on \textit{2D Darcy}, $\sigma$ denotes the predictive standard deviation (STD). The last two columns show the point-wise predictive std normalized by the ground-truth.}
%	\label{fig:darcy_plot}
%\end{figure}
%\subsection{Uncertainty Quantification}
\noindent\textbf{Uncertainty Calibration}. Next, we evaluated our method from a probabilistic perspective by examining the Mean Negative Log-Likelihood (MNLL). For comparison, we trained FNO and GNOT using two widely adopted Bayesian neural network training approaches: Stochastic Gradient Langevin Dynamics (SGLD)~\citep{welling2011bayesian} and Monte Carlo Dropout (MCDropout)~\citep{gal2016dropout}. For SGLD, the initial learning rate was chosen from the range $\{10^{-7}, 10^{-6}, 10^{-5}, \ldots, 10^{-2}\}$, while for MCDropout, the dropout rate was tuned across $\{0.1, 0.2, \ldots, 0.5\}$. \cmt{The resulting MNLL values and standard deviations are reported in Table~\ref{tab:nll}.} As shown in Table~\ref{tab:nll}, both variants of our method ---\ours-FT and \ours-QR--- consistently achieve the lowest MNLL across all the datasets, largely outperforming the competing methods. These results highlight the strength of our method not only in predictive accuracy but also in uncertainty calibration.

	We further investigated the probabilistic predictions of our method. Specifically, we randomly selected four test examples from two real-world applications --- \textit{Beijing-Air} and \textit{Quasar} --- as well as six test examples from the simulated applications. Figure~\ref{fig:model-pred-beijing} and Appendix Figure~\ref{fig:model-pred-quasar} present the predictive means and standard deviations produced by \ours-FT and \ours-QR at each input location for the real-world datasets. As shown, in regions where the predictive mean closely matches the ground truth, the shaded region --- representing the predictive standard deviation (STD) --- is relatively narrow. Conversely, when the discrepancy between the predictive mean and the ground truth increases, the shaded region widens, indicating higher predictive uncertainty. Notably, the predictive standard deviation produced by \ours-FT appears smoother than that of \ours-QR. This behavior may be attributed to the fact that \ours-QR freely learns the weight function at the quadrature nodes, whereas \ours-FT leverages Fourier transforms that primarily retain low-frequency components, resulting in smoother uncertainty estimates.
	Appendix Figure~\ref{fig:darcy_plot} further compares predictive STDs with point-wise errors and their normalized counterparts on test examples from the \textit{Darcy} dataset. In both absolute and normalized cases, larger errors generally correspond to larger predictive STDs. After normalization, however, regions with large absolute errors often exhibit smaller relative errors and consequently smaller normalized STDs. This suggests that the predictive STD naturally scales with the magnitude of the ground-truth values, which is a desirable property.
	In addition, Appendix Figure~\ref{fig:car_plot} shows that, for the \textit{Car Shape} dataset, the predictive STD consistently aligns with the error of the predictive mean across the selected examples. Finally, as illustrated in Appendix Figure~\ref{fig:burgers_plot}, the \textit{Burgers} dataset exhibits extremely small prediction errors (NMSE: $0.005$), which correspond to correspondingly small predictive STDs.
	Overall, the standard deviation outputs from our method appropriately reflect the quality of the predictions, confirming that \ours provides well-calibrated uncertainty estimates.

%We further investigated the probabilistic predictions of our method. Specifically, we randomly selected four test examples from the \textit{Beijing-Air} dataset and another four from the \textit{Quasar} dataset. Figure~\ref{fig:model-pred-beijing} and Appendix Figure~\ref{fig:model-pred-quasar} display the predictive means and standard deviations produced by \ours-FT and \ours-QR at each input location. As observed, in regions where the predicted mean closely matches the ground truth, the shaded region---representing the predictive standard deviation---is relatively narrow. Conversely, in regions where the discrepancy between the predictive mean and the ground truth is larger, the shaded area expands, indicating higher predictive uncertainty. It is interesting to observe that the predictive standard deviation produced by \ours-FT is smoother than that of \ours-QR. This may be because \ours-QR freely learns the weight function at the quadrature nodes, whereas \ours-FT leverages Fourier transforms to retain primarily low-frequency components, resulting in smoother uncertainty estimates.
%Overall, the standard deviation outputs from our method appropriately reflect the quality of the predictions, confirming that \ours provides well-calibrated uncertainty estimates.

\noindent\textbf{Ablation Studies.} We conducted extensive ablation studies to further evaluate \ours, examining various hyperparameter choices, the number of projection points,  integral transforms, and the effectiveness of GP activation function. We also analyzed the learned weight functions, and evaluated the training efficiency.  Details are provided in Appendix Section~\ref{sect:ablation}, ~\ref{sect:weight}, and~\ref{sect:time}.

\vspace{-0.05in}
\section{Conclusion}
\vspace{-0.05in}
We have presented \ours, a deep Gaussian process model designed for learning mappings between functions. The model comprises a sequence of GP layers that perform linear transformations and nonlinear activations in functional space. \ours is capable of handling sparse, noisy, and arbitrarily irregularly sampled data, while providing principled  probabilistic inference for effective uncertainty calibration. Its performance on both simulated and real-world datasets is encouraging, outperforming or performing on par with classical functional linear regression and recent neural operator methods.

\section*{Impact Statement}
This paper presents work whose goal is to advance the field of Machine
Learning. There are many potential societal consequences of our work, none
which we feel must be specifically highlighted here.

\bibliographystyle{icml2026}
\bibliography{references}

%%%%%%%%%%%%%%%%%%%%%%%%%%%%%%%%%%%%%%%%%%%%%%%%%%%%%%%%%%%%%%%%%%%%%%%%%%%%%%%
%%%%%%%%%%%%%%%%%%%%%%%%%%%%%%%%%%%%%%%%%%%%%%%%%%%%%%%%%%%%%%%%%%%%%%%%%%%%%%%
% APPENDIX
%%%%%%%%%%%%%%%%%%%%%%%%%%%%%%%%%%%%%%%%%%%%%%%%%%%%%%%%%%%%%%%%%%%%%%%%%%%%%%%
%%%%%%%%%%%%%%%%%%%%%%%%%%%%%%%%%%%%%%%%%%%%%%%%%%%%%%%%%%%%%%%%%%%%%%%%%%%%%%%
\newpage
\appendix
\onecolumn
\section*{Appendix}
\section{GP Covariance for Integral Transformation}\label{sect:cross-cov}
Suppose a stochastic function $f$ is sampled from a GP prior with covariance function as a kernel function $\kappa(\cdot, \cdot)$, 
\[
f \sim \gp(0, \kappa(\x, \x')).
\]
From the weight space view~\citep{Rasmussen06GP}, we can represent: $$f(\x) = \phi(\x)^\top \w,$$ where $\phi(\x)$ is the implicit feature mapping of the kernel, \ie $\kappa(\x, \x') = \bphi(\x)^\top \bphi(\x')$, and $\w \sim \N(0, \I)$. Note that the feature mapping $\phi(\cdot)$ can be infinitely dimensional. 
 Suppose function $h$ is a linear transformation of $f$, 
\begin{align}
	h(\x) = \int_\Omega  \Wcal(\x, \z)f(\z) \d \z =  \int_\Omega \Wcal(\x, \z) \bphi(\z)^\top \w \d \z,
\end{align}
where $\Wcal$ is the coefficient (or weight) function.  We can accordingly compute the cross-covariance function between $h$ and $f$ at any pair of inputs $(\x, \x')$ by
\begin{align}
	&\cov(h(\x), f(\x')) = \EE\left[h(\x)f(\x')\right] - \EE[h(\x)]\EE[f(\x')] \notag \\
	&= \int_\Omega \Wcal(\x, \z)\bphi(\z)^\top \EE\left[\w \w^\top\right] \bphi(\x') \d \z  \notag \\
	&=\int_\Omega \Wcal(\x, \z)\bphi(\z)^\top \bphi(\x') \d \z  \notag \\
	&=\int_\Omega \Wcal(\x, \z)\kappa(\z, \x') \d \z. \label{eq:cross}
\end{align}
Similarly, we can derive the covariance function of $h$:
\begin{align}
	\cov(h(\x), h(\x')) &= \int_\Omega\int_\Omega \Wcal(\x, \z) \kappa(\z, \z') \Wcal(\z', \x') \d\z \d \z'. \label{eq:self}
\end{align}
From \eqref{eq:cross} and \eqref{eq:self} we can see that these (cross-) covariance functions are non-stationary even if both $\Wcal$ and $\kappa$ are stationary, namely when $\Wcal(\x, \z) = \Wcal(\x-\z)$ and $\kappa(\z, \x')=\kappa(\z - \x')$.
\section{Model Details}\label{section:model-details}
\subsection{Regarding Prior~\eqref{eq:prior-over-f}}
Let $\f^j$, $\f^j_Q$, and $\widehat{\f}^j$ denote the values of $f_j$ at $\X_{\text{in}}$, $\X_Q$, and the noisy observations at $\X_{\text{in}}$, respectively. Define $\widehat{\F} = [\widehat{\f}^1,\ldots,\widehat{\f}^{d_0}]$, $\F = [\f^1,\ldots,\f^{d_0}]$, and $\F_Q = [\f^1_Q,\ldots,\f^{d_0}_Q]$. Their joint distribution factorizes as	$$p(\widehat{\F}, \F, \F_Q) = p(\F)p(\widehat{\F}|\F) p(\F_Q|\F) = \prod\nolimits_{j=1}^{d_0}\N(\f^j|\0, \nu_j(\X_{\text{in}}, \X_{\text{in}})) \N(\widehat{\f}^j|\f^j, \sigma^2_j\I) p(\f^j_Q|\f_j),$$ where $\sigma^2_j$ is the noise variance, and $p(\f^j_Q|\f_j)$ is conditional Gaussian. Marginalizing  out $\F$ yields the prior distribution in~\eqref{eq:prior-over-f}.
\subsection{GP Priors in Functional Space}\label{sect:cond-gp-priors}
Our model constructs a sequence of conditional GP prior in the functional space. Specifically, at each layer $l$, When $h_{l,i} \rightarrow h_{l+1,i}$ is a  nonlinear transformation (see~\eqref{eq:nonlinear-trans}), it implies 
\begin{align}
	h_{l+1, i}(\cdot)\mid h_{l,i}(\cdot) \sim \mathcal{GP}, 
\end{align}
with covariance function, 
\begin{align}
	&\text{cov}\left(h_{l+1, i}(\x), h_{l+1, i}(\x') \mid h_{l, i}(\cdot)\right) \notag \\
	%&= \vartheta_l\left(h_{l,i}(\x), h_{l,i}(\x')\right) - 
	%\vartheta_l\left(h_{l,i}(\x), \bbeta\right)\vartheta_l\left(\bbeta, \bbeta\right)^{-1} \vartheta_l\left(\bbeta, h_{l,i}(\x')\right).
	&= 
	\vartheta_l\left(h_{l,i}(\x), \bbeta\right)\vartheta_l\left(\bbeta, \bbeta\right)^{-1} \vartheta_l\left(\bbeta, h_{l,i}(\x')\right).
\end{align}
%where $t_l(\x) = \vartheta_l\left(h_{l,i}(\x), h_{l,i}(\x)\right) - \vartheta_l\left(h_{l,i}(\x), \bbeta\right)\vartheta_l(\bbeta, \bbeta)^{-1}\vartheta_l\left(\bbeta, h_{l,i}(\x)\right)$.

When $h_{l,i} \rightarrow h_{l+1,i}$ is a linear transformation as in~\eqref{eq:actual-linear-transform}, it  induces a conditional GP prior:
\begin{align}
	h_{l+1,i}(\cdot) \mid h_{l-1,i}(\cdot) \sim \mathcal{GP},
\end{align}
with covariance function,
\begin{align}
	&\text{cov}(h_{l+1,i}(\x), h_{l+1,i}(\x') \mid h_{l-1,i}(\cdot)) \notag \\
	&= c_{l, i}(\x, \X_Q) k_{l,i}(\X_Q, \X_Q)^{-1} \text{cov}\left(\h_{l,i} \mid h_{l-1,i}(\cdot)\right) k_{l,i}(\X_Q, \X_Q)^{-1} c_{l, i}(\X_Q, \x').
\end{align}
\section{Dataset Details}\label{sect:dataset}
\subsection{Burgers}
We used the following Burger's equation:
\begin{align}
	u_t + u_{xx} = \nu u_{xx}, u(x, 0) = u_0(x), \label{eq:burgers}
\end{align}
where $(x, t) \in [0, 1]^2$, and $u_0(x)$ is the initial condition, and $\nu=0.1$ is the viscosity. We aim to learn a mapping from the initial condition to the solution at $t=1$, namely, $u_0 \rightarrow u_1(x) \overset{\Delta}{=}u(x, 1)$. The initial condition $u_0$ is sampled from a Gauss random field, $\N(0, 625(-\Delta + 25\I)^{-2})$. The dataset was generated in~\citep{lu2022comprehensive}, with each pair of input and output functions sampled at the same set of 128 equally spaced locations across the spatial domain. 

\subsection{Darcy Flow}
We  employed a Darcy flow equation in a rectangle domain: 
\begin{align}
	-\nabla (c(\x) \nabla u(\x)) =1,  \label{eq:darcy}
\end{align}
where $\x \in [0, 1]^2$,  $c(\x)>0$ is the permeability field,  and $u(\x) = 0$ at the boundary. The goal is to predict the solution field from the permeability field:  $c \rightarrow u$. The permeability field $c$ a piece-wise constant function derived by first sampling  a continuous function from a Gauss random field $\N(0, (-\Delta + 9\I)^2)$, and then mapping the positive values to 12 and the negative values to 3. Every input-output function pair is discretized on a $29 \times 29$ uniform grid over the input domain. The dataset was generated and shared by~\citet{lu2022comprehensive}.
\subsection{Car Shape}
The dataset was generated by~\citet{umetani2018learning}, where the Reynolds-averaged Navier–Stokes (RANS) equations, coupled with a turbulence model and SUPG stabilization, are solved using a finite element method to compute time-averaged velocity and pressure fields. The inlet velocity is fixed at 20 m/s, corresponding to an estimated Reynolds number of $5 \times 10^6$.  The car geometries are drawn from the ShapeNet car category~\cite{chang2015shapenet}, with each car surface discretized into approximately 3,700 mesh points.

%\subsection{3D Compressible Naiver-Stoke (NS) Equations }
%The third scenario involves 3D compressible NS equations:
%\begin{align}
%	\partial_t \rho + \nabla \cdot (\rho \v) &= 0, \notag \\
%	\rho\left(\partial_t \v + \v \cdot \nabla \v\right) &= -\nabla p + \eta \Delta \v + (\zeta + \eta/3) \nabla(\nabla \cdot \v), \notag \\
%	\partial_t \left[ \epsilon + \frac{\rho v^2}{2}\right] &+ \nabla \cdot \left[\left(\epsilon + p + \frac{\rho v^2}{2}\right)\v - \v \cdot \sigma'\right]= 0,  
%\end{align}
%where $\rho$ is the mass density, $\v$ is the velocity, $p$ is the gas pressure, $\epsilon=p/(\Gamma - 1)$ is the internal energy, $\Gamma=5/3$, $\sigma'$ is the viscous stress tensor, and $\eta, \zeta$ are the shear and bulk viscosity, respectively.  The behavior of the fluid is sensitive to the Mach number $M = |v|/c_s$, where $c_s = \sqrt{\Gamma p/\rho}$. The data were generated and made available through PDEBench~\citep{takamoto2022pdebench}, a widely used benchmark dataset for scientific machine learning. We considered the high Mach number case ($M = 1.0$), where the fluid behavior is complex, making the learning task challenging. The input and output functions are discretized on a uniform grid of size $64 \times 64 \times 64$.

\begin{figure*}%[h]
	\centering
	\includegraphics[width=\linewidth]{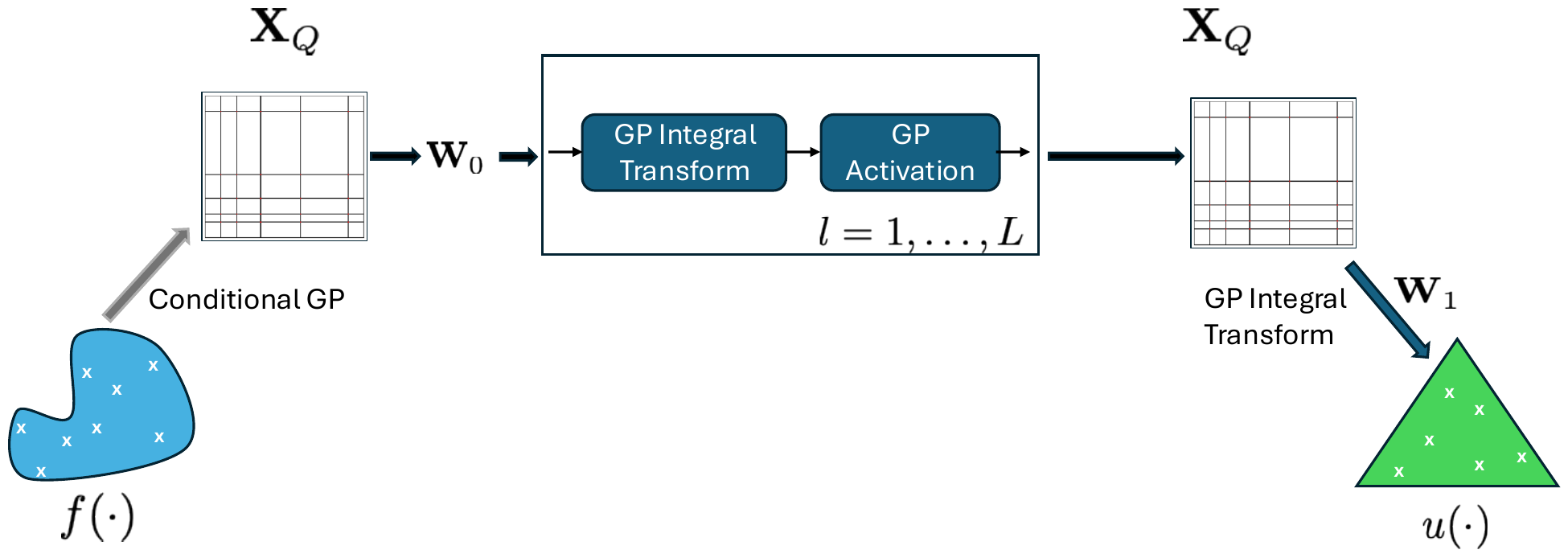}
	\caption{\small Deep Gaussian processes for functional maps (\ours).} 
	\label{fig:model-arch}
\end{figure*}

\subsection{Quasar Reverberation Mapping}\label{sect:lights}
In astronomy, understanding the relationship between the central continuum emission of a quasar and the subsequent response from surrounding emitting regions is key to inferring its physical properties, structure, and kinematics --- a process known as reverberation mapping~\citep{1982ApJ...255..419B, 1993PASP..105..247P}. This task is often modeled via an unknown transfer function that links the response emission to the driving continuum emission, naturally framing the problem as function-on-function regression.

The Zwicky Transient Facility (ZTF)~\citep{2019PASP..131a8002B}, located at the Palomar Observatory, is an automated time-domain survey utilizing a 4-foot Schmidt telescope equipped with a 47.2-square-degree field-of-view camera. ZTF scans the entire Northern sky with a cadence of approximately three nights during Phase I (May 2018–September 2020) and two nights during Phase II (December 2020–present) for its custom $g$-band and $r$-band photometric filters, with a four-night cadence for the $i$-band.

To construct a dataset aligned with this  task, we collected $g$-band and $r$-band light curves from the most recent ZTF data release (DR23)~\citep{Masci_2018}, focusing on the first 18,000 objects in the Million Quasars catalogue~\citep{Flesch_2023}. In this setting, we treat the shorter-wavelength $g$-band light curve as the input function driving the response observed in the $r$-band light curve, which serves as the output function.

We preprocessed the raw data following the methodology of~\citep{S_nchez_S_ez_2021}. Specifically, we retained light curves with mean magnitudes slightly brighter than the ZTF limiting magnitude of 20.6 and fainter than 13.5 to avoid saturated measurements. We excluded observations with magnitude errors exceeding one and with non-zero catflags quality scores. However, we did not filter light curves based on variability features.

We further restricted the data to observations within the first 2,000 days and randomly sampled up to 500 time points from each light curve, provided sufficient data existed. This resulted in 793 pairs of irregularly sampled light curves, with differing time points across the input and output functions for each example---thereby offering a suitable testbed for function-on-function regression. For our experiments, we randomly split the dataset into 650 training and 143 testing examples.
%\begin{figure*}%[h]
%	\centering
%	\includegraphics[width=\linewidth]{figs/model-architecture-trim.pdf}
%	\caption{\small Model framework illustration}. 
%	\label{fig:model-arch}
%\end{figure*}
\section{Hyperparameter Selection}\label{sect:hyper}
Here we provide hyperparameter selection details for each method.
\begin{itemize}
	\item \textbf{FLR}: We adopted the implementation from the Scikit-FDA library\footnote{\url{https://fda.readthedocs.io/en/latest/}} for LFR-Fourier and LFR-BSpline. The primary hyperparameter is the number of bases, which was selected from \{2, 3, \ldots, 30\}. The range of each basis function is set to a minimum range that covers the observed output function values, \eg [0, 1] for \textit{Burgers} and \textit{Darcy}. The intercept parameter was jointly estimated with the basis coefficients. We employ the second order differential operator regularization, which is the default choice of the library. The implementation of FLR-GP is directly from that of \ours. The selection of the hyperparameters is shared with that for \ours, except we fixed the number of GP layers to one, and no GP activation is invovled. 
	\item \textbf{FNO}\footnote{\url{https://github.com/neuraloperator/neuraloperator}}:  The hyperparameters include the number of modes, which varies from \{8, 10, 12, 16, 20\}, the number of channels for channel lifting, which varies from \{8, 16, 32, 64, 128, 256\}, and the number of Fourier layers, which varies from \{2, 3, 4\}. We used GELU activation, the default choice in the official  library. 
	\item \textbf{GNOT}\footnote{\url{https://github.com/HaoZhongkai/GNOT}}: the hyperparameters include the number of attention layers, varying from \{3, 4, 5\}, the dimensions of the embeddings, varying from \{8, 16, 32, 64\}, and the inclusion of mixture-of-expert-based gating, specified as either \{yes, no\}. We used GeLU activation, the default choice of the official library.
%	\item F-FNO. : The set of hyperparameters are the same as FNO, including the number of modes varying from \{8, 10, 12, 16, 20\}, the number of latent channels tuned from \{8, 16, 32, 64\}, and the number of one-dimensional integration layers, which varies from \{2, 3, 4\}. We used ReLU activation, the default choice in the original library\footnote{\url{https://github.com/alasdairtran/fourierflow}}. We used a simple linear mapping that maps the irregular locations at each dimension to a evenly-spaced grid in [0, 1]~\citep{li2023fourierGEO}.
	\item \textbf{DSE-FNO}\footnote{\url{https://github.com/camlab-ethz/DSE-for-NeuralOperators}}: The set of hyperparameters are the same as FNO,  including the number of modes chosen from \{8, 10, 12, 16, 20\}, the number of latent channels  from \{8, 16, 32, 64, 128, 256\}, and the number of integration layers from \{2, 3, 4\}.  The activation was chosen from \{GeLU, ReLU, SiLU\}.
	\item \textbf{\ours}: Our method was implemented using JAX~\citep{frostig2018compiling}. We used the ADAM optimizer, with the initial learning rate selected from \{5e-5, 1e-4, 5e-4\}, and a cyclical cosine annealing schedule with the max learning rate as $0.001$.  The number of training epochs was chosen from \{100, 250, 500, 1000, 5000, 10000\}. 
	The number of GP layers was varied from \{2, 3, 4\}. The covariance functions for the input function, weight functions, and the GP activation, were selected from the Square Exponential (SE) kernel, Mat\'ern kernel with degree of freedom $5/2$ and $7/2$, or a weighted combination of two Mat\'ern kernels with degrees of freedom $5/2$  and $13/2$. The number of inducing points for each GP activation was selected from $\{32, 64, 128, 256, 512\}$, and the column dimension of the weight matrices $\W_0$ and $\W_1$ from $\{4, 8, 16, 32, 64, 128, 256\}$. For \ours-FT on datasets sampled at regular grids, we kept the number of projection locations the same as the number of locations in the original functions, and on the irregular sampled datasets searched over \{32, 64, \ldots, 512\} (using their tensor product for higher dimensional problems). In a similar capacity, for \ours-QR we used a trapezoidal quadrature rule at the function locations when handling problems on a regular grid, and for the irregular grid problems, used a tensor-product Gauss-Legendre rule with the number of one-dimensional nodes as selected from \{32, 64, \ldots, 512\}. For the Car Shape dataset,  we used vertex-based quadrature rules with barycentric areas defined over the trangular meshes. 
	\end{itemize}
All the neural operator methods (FNO, GNOT, DSE-FNO) used ADAM optimization with learning rate selected from $\{10^{-5}, 5\times 10^{-5}, 10^{-4}, 4\times 10^{-3}, 10^{-3}\}$. The maximum number of epochs was set to 10000, which ensures convergence. The batch size was set to 500 for \textit{Beijing-Air} dataset and 100 for all the other datasets.  We ran all the methods on NCSA Delta GPU cluster\footnote{\url{https://www.ncsa.illinois.edu/research/project-highlights/delta/}}, with NAVIDA A40 GPUs.

\begin{figure*}%[h]
	\centering
	\includegraphics[width=\linewidth]{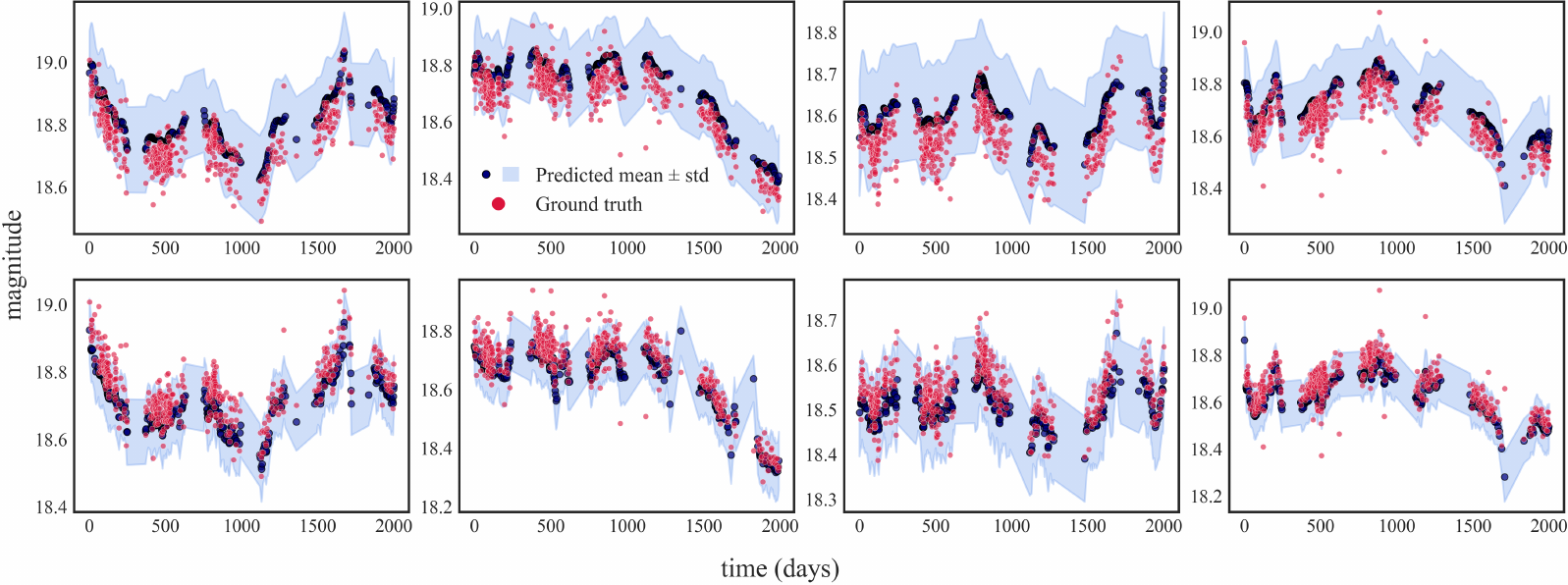}
	\caption{\small Prediction examples of \ours on \textit{Quasar} dataset. The shaded regions indicate one predictive standard deviation. The top row shows the prediction of \ours-FT and the bottom row \ours-QR.  }
	\label{fig:model-pred-quasar}
\end{figure*}
\begin{figure*}
	\centering
	\includegraphics[width=1.\linewidth]{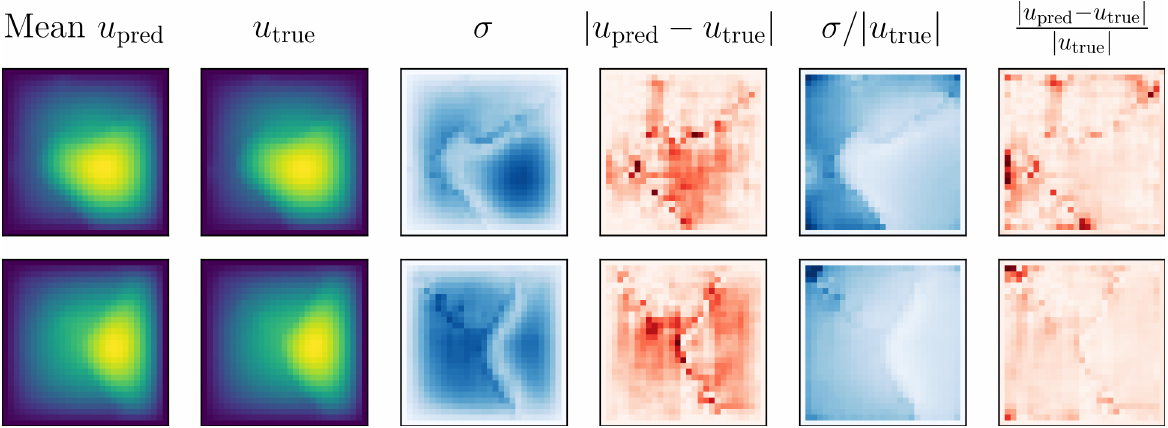}
	\caption{\small Prediction examples of \ours-QR on \textit{Darcy}, $\sigma$ denotes the predictive standard deviation (STD). The last two columns show the point-wise predictive std normalized by the ground-truth.}
	\label{fig:darcy_plot}
\end{figure*}
\begin{figure*}
	\centering
	\includegraphics[width=0.8\linewidth]{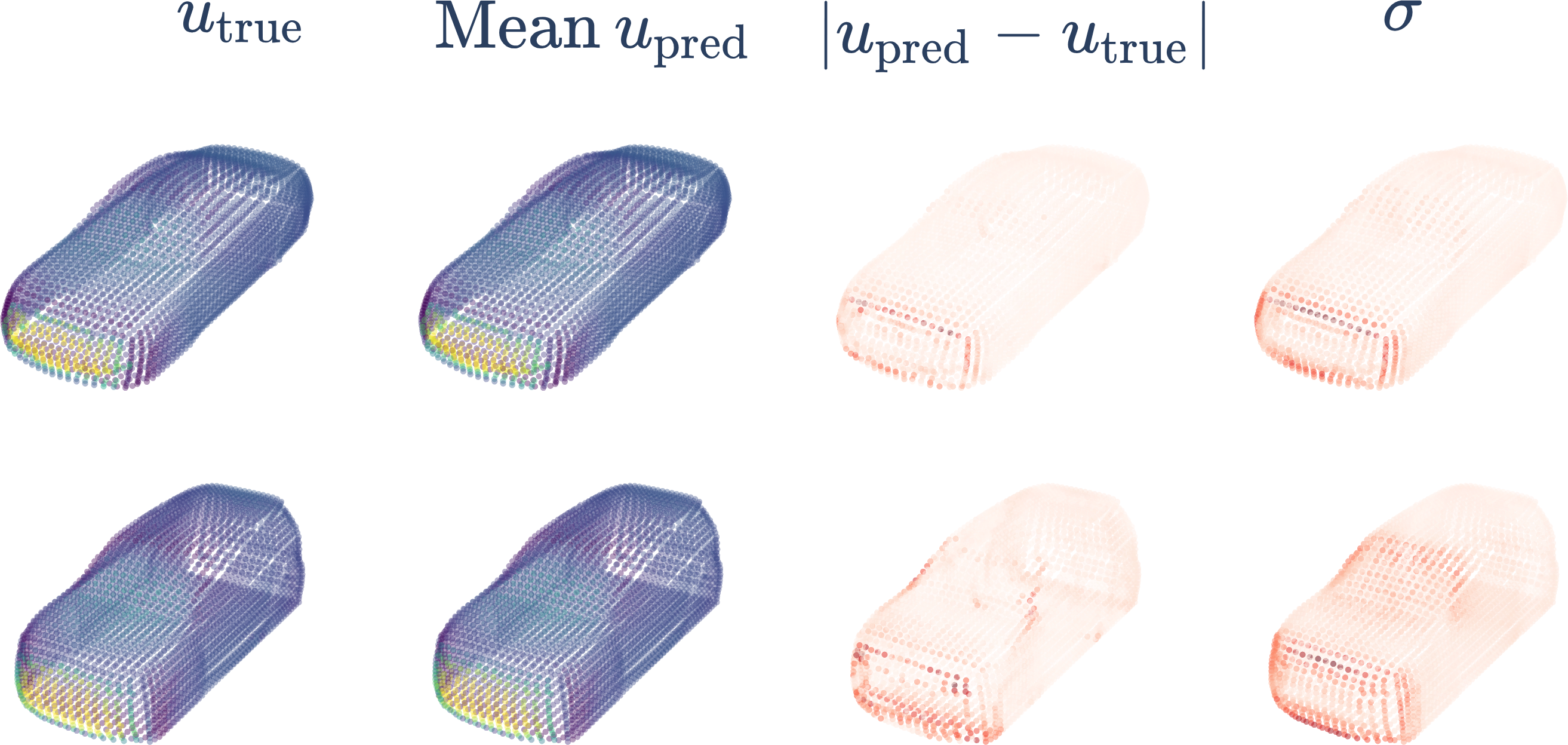}
	\caption{\small Prediction examples of \ours-QR on \textit{Car Shape}, $\sigma$ denotes the predictive standard deviation (STD).}
	\label{fig:car_plot}
\end{figure*}
\begin{figure*}
	\centering
	\includegraphics[angle=270, width=0.9\linewidth]{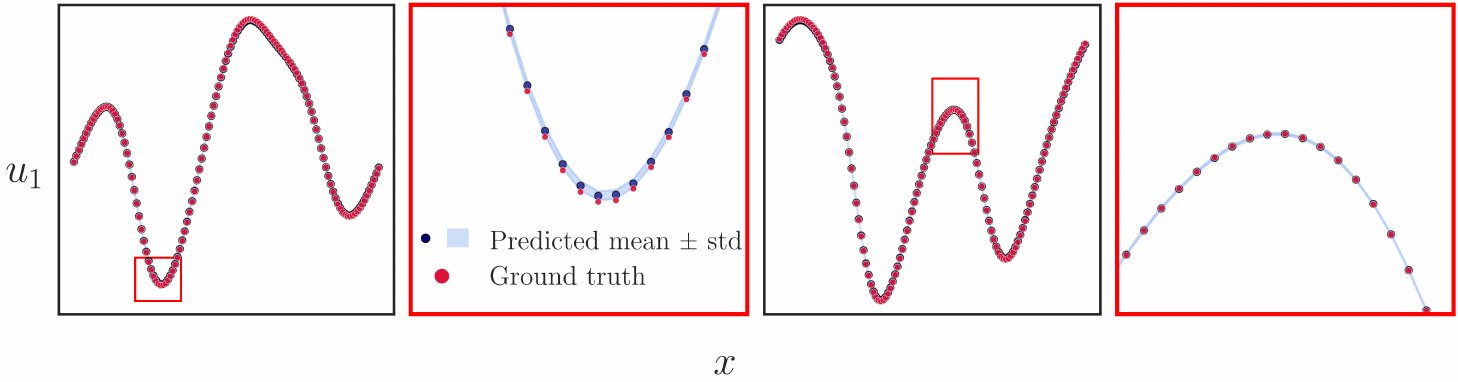}
	\caption{Prediction examples of  \ours-FT on \textit{Burgers}.}
	\label{fig:burgers_plot}
\end{figure*}

\section{Error Analysis of Discrete Integral Transform}\label{sect:error-analysis}
To perform discrete integral transforms, we used several classical numerical approximation methods:  Gauss-Legendre (GL) Qudrature, Trapezoidal (TR) Rules, and Fourier-based Convolution. Their theoretical error properties are well established and summarized as follows.  Specifically, for a given function $f$, 
\begin{itemize}
	\item \textbf{Gauss-Legendre Quadrature}:
	\begin{itemize}
		\item An $N$-point GL rule is exact for polynomials up to degree $2N-1$.
		\item For $f\in C^k$, error is $O(1/N^k)$ (no periodicity constraint for this like in Trapezoidal rule); for analytic $f$, error is $O(e^{-cN})$.
	\end{itemize}
	\item \textbf{Trapezoidal Rule}:
	\begin{itemize}
		\item On $[a,b]$, the error is $O(1/N^2)$ if $f''$ exists and is continuous; degrades to $O(1/N)$ if $f\in C^0$.
		\item For periodic $f\in C^m$, error improves to $O(1/N^m)$; analytic periodic functions yield exponential convergence.
	\end{itemize}
	\item \textbf{Fourier-based Convolution}:
	\begin{itemize}
		\item If $f$ is periodic and $C^m$, FFT convolution matches TR’s periodic rates: $O(1/N^m)$ or exponential if analytic.
		\item If $f$ is non-periodic, the implicit periodic extension introduces boundary jumps, causing error stuck at $O(1/N)$ regardless of interior smoothness.
	\end{itemize}
\end{itemize}

Our model applies  a dimension-wise integral transform instead of performing a full-dimensional one. If we consider the full-dimensional transform as the ground truth, the approximation error introduced by our dimension-wise approach can be expressed (in the 2D case) as:
\begin{align}
\Delta_1 =  \bigl((\mathcal{T}-\widetilde{\mathcal{T}})h\bigr)(x_1,x_2) = \iint \left\{w(x_1,x_2,y_1,y_2) - \left[w_1(x_1,y_1) + w_2(x_2,y_2)\right]\right\} h(y_1,y_2) \, dy_1 \, dy_2. \notag 
\end{align}
If the full weight function $w(\cdot)$ can be decomposed as the sum of independent components $w_1$ and $w_2$, the error vanishes, \ie $\Delta_1 = 0$.  Otherwise, a gap remains. However, since $w(\cdot)$, $w_1$, and $w_2$ are unknown a priori and must be learned from data, the approximation error is inherently data-dependent.

From a modeling perspective, our approach adopts a reduced model space (or equivalently, a simpler inductive bias) to represent the weight function(s) in the integral transform, which is \textit{not} necessarily a limitation. Instead, the dimension-wise transform enables the model's parameter complexity to scale linearly with input dimensionality. In contrast, using a full-dimensional transform causes exponential growth in the number of parameters needed to learn the weight function, significantly increasing training cost and the risk of overfitting. Our ablation study has confirmed this point. See Table~\ref{tab:dimwise-ablation}. 

The total error regarding our discrete dimension-wise integral transform can be decomposed as two parts. The first one is the aforementioned $\Delta_1$. The second part comes from the numerical approximation error. 
Each of the dimension-wise integrals (\eg those involving $w_1$ and $w_2$) is then approximated using numerical quadrature or discrete Fourier transforms, introducing a separate approximation error, denoted as $\Delta_2$. The analysis of $\Delta_2$ has already been provided. The total error can be expressed as $\Delta = |\Delta_1| +| \Delta_2|$.

%
%Similarly, as shown in Eq. 9, each $\hat{f}_j(x)$ is marginally assigned a GP prior. Our model then alternates between defining conditional GP priors across layers, ultimately generating the output function $u$. Finally, we use a Gaussian likelihood $p(Y \mid U)$ to model the data fit. 
%depth, mode, channels, inducing point, kernel choice
\section{Ablation Studies}\label{sect:ablation}
We conducted a series of ablation studies to further evaluate our method.  

\noindent\textbf{Hyperparameters.} 
We performed two comprehensive ablation studies --- one on the real-world dataset \textit{Beijing-Air} and the other on the simulated dataset \textit{Darcy} --- to investigate the influence of hyperparameter choices. For \textit{Beijing-Air}, we examined all major DGPFM-FT hyperparameters, including the number of integration (linear) layers, Fourier modes, latent channels ($C$), inducing points ($S$), and kernel/covariance choices. The base model consisted of 4 integration layers, 10 Fourier modes, 256 channels, 32 inducing points, and a weighted Matérn kernel (DOF 5/2 and 13/2). We varied each hyperparameter independently while fixing the others and evaluated performance using normalized root mean square error (NRMSE) and negative log-likelihood (NLL). To ensure statistical reliability, each configuration was run five times, and we report mean NRMSE and NLL values.
The second study ablated DGPFM-QR hyperparameters on the \textit{Darcy Flow} dataset, with a base model of 5 integration layers, 64 latent channels, 64 inducing points, and the same kernel setup. Results are summarized in Table~\ref{tab:dgpfm-ft-ablation} and Table~\ref{tab:dgpfm-qr-ablation}.

The ablations show that DGPFM-FT’s expressivity improves with additional integration layers, channels, and Fourier modes, as reflected in NRMSE. However, beyond four integration layers or larger model sizes, NLL improvements diminish --- likely due to the increasing difficulty of variational inference optimization. Varying the number of inducing points produced no clear trend, suggesting this hyperparameter should be tuned per dataset. Mat\'ern kernels consistently outperformed the Squared Exponential, indicating that kernel smoothness has a significant impact on performance.

For \ours-QR, performance improved with more layers, channels, and inducing points, but NLL began to degrade once the number of inducing points exceeded 16. As with DGPFM-FT, finitely smooth kernels were advantageous; interestingly, the Mat\'ern 5/2 kernel outperformed the 13/2 variant in NLL, while yielding higher NRMSE --- the opposite of the trend observed on \textit{Beijing-Air}. 
This dataset-dependent discrepancy motivated our use of weighted combinations of Mat\'ern kernels, enabling the model to adaptively learn an appropriate level of smoothness.
%As before, increasing model size improved NRMSE but yielded diminishing returns in NLL beyond a certain point.

\begin{table}[t]
	\small
	\centering
	\caption{\small DGPFM-FT ablations on \textit{Beijing-Air}. The base model uses 4 integration layers, 10 Fourier modes, 256 channels, 32 inducing points, and a weighted Matern kernel (DOF 5/2 and 13/2). Best results are shown in bold. }\label{tab:dgpfm-ft-ablation}
	%\scriptsize
	\begin{subtable}{\textwidth}
		\centering
		\caption{\small  The number of integration layers.}
		\begin{tabular}{ccccccc}
			\toprule
			\#\textbf{Int Layers}  & 1 & 2 & 3 & 4 & 5 & 6 \\
			\midrule
			NRMSE & 0.583 & 0.539 & 0.373 & 0.288 & 0.263 & \textbf{0.253} \\
			NLL   & N/A   & 21.97 & 7.84  & \textbf{7.80} & 8.12 & 8.37 \\
			\bottomrule
		\end{tabular}
	\end{subtable}
	\vspace{0.1in}
	
	\begin{subtable}{\textwidth}
		\centering
		\caption{\small The number of latent channels $C$.}
		\begin{tabular}{cccccccc}
			\toprule
			\textbf{Channels} ($C$) & 8 & 16 & 32 & 64 & 128 & 256 & 512 \\
			\midrule
			NRMSE & 0.521 & 0.491 & 0.462 & 0.411 & 0.375 & 0.288 & \textbf{0.242} \\
			NLL   & 9.40  & 8.47  & 8.45  & 8.24  & 8.40  & \textbf{7.80}  & 8.78\\
			\bottomrule
		\end{tabular}
	\end{subtable}
	\vspace{0.1in}
	
	\begin{subtable}{\textwidth}
		\centering
		\caption{\small The number of inducing points $S$.}
		\begin{tabular}{ccccccc}
			\toprule
			\textbf{Inducing Points ($S$)} & 4 & 8 & 16 & 32 & 64 & 128 \\
			\midrule
			NRMSE & 0.271 & 0.311 & \textbf{0.269} & 0.288 & 0.270 & 0.273 \\
			NLL   & 20.04 & 8.99  & 8.73  & \textbf{7.80} & 8.54 & 9.09 \\
			\bottomrule
		\end{tabular}
	\end{subtable}
	\vspace{0.1in}
	\begin{subtable}{\textwidth}
		\centering
		\vspace{0.1in}
		\caption{\small Number of Fourier modes.}
		\begin{tabular}{ccccccc}
			\toprule
			\textbf{Modes} & 4 & 8 & 12 & 16 & 32 & 64 \\
			\midrule
			NRMSE & 0.391 & 0.305 & 0.277 & 0.255 & 0.224 & \textbf{0.219} \\
			NLL   & 7.94  & 8.38  & 7.91  & 8.13  & \textbf{7.81} & 8.30 \\
			\bottomrule
		\end{tabular}
	\end{subtable}
	%\vspace{0.1in}
	\begin{subtable}{\textwidth}
		\centering
			\vspace{0.1in}
		\caption{\small Choice of kernels.}
		\begin{tabular}{ccccccc}
			\toprule
			\textbf{GP Kernel} & Squared Exp & Matérn 5/2 & Matérn 13/2 & \multicolumn{3}{c}{Weighted Matérn (5/2+13/2)} \\
			\midrule
			NRMSE & 0.321 & 0.289 & 0.308 & \multicolumn{3}{c}{\textbf{0.288}} \\
			NLL   & 8.26  & 8.26  & 8.21  & \multicolumn{3}{c}{\textbf{7.80}} \\
			\bottomrule
		\end{tabular}
	\end{subtable}
\end{table}

\begin{table}[t]
	\small
	\centering
	\caption{\small DGPFM-QR ablations on \textit{Darcy-Flow}. The base model uses 5 integration layers, 64 channels, 64 inducing points, and a weighted Matern kernel (DOF 5/2 and 13/2). Best results are shown in bold. }\label{tab:dgpfm-qr-ablation}
	%\scriptsize
	\begin{subtable}{\textwidth}
		\centering
		\caption{\small The number of integration layers.}
		\begin{tabular}{ccccccc}
			\toprule
			\#\textbf{Int Layers} & 1 & 2 & 3 & 4 & 5 & 6 \\
			\midrule
			NRMSE & 9.97e-2 & 2.67e-2 & 2.27e-2 & 1.97e-2 & 1.86e-2 & \textbf{1.80e-2} \\
			NLL   & N/A   & 2.04 & -2.97  & -3.71  & -4.06 & \textbf{-4.13} \\
			\bottomrule
		\end{tabular}
	\end{subtable}
	\vspace{0.1in}
	
	\begin{subtable}{\textwidth}
		\centering
		\caption{\small The number of latent channels $C$.}
		\begin{tabular}{ccccccc}
			\toprule
			\textbf{Channels} ($C$) & 8 & 16 & 32 & 64 & 128 & 256 \\
			\midrule
			NRMSE & 2.47e-2 & 2.29e-2 & 1.99e-2 & 1.86e-2 & \textbf{1.72e-2} & 1.89e-2 \\
			NLL   & -2.46  & -3.64  & -3.94  & \textbf{-4.06}  & -3.07  & -3.61 \\
			\bottomrule
		\end{tabular}
	\end{subtable}
	\vspace{0.1in}
	
	\begin{subtable}{\textwidth}
		\centering
		\caption{\small The number of inducing points $S$.}
		\begin{tabular}{ccccccc}
			\toprule
			\textbf{Inducing Points ($S$)} & 4 & 8 & 16 & 32 & 64 & 128 \\
			\midrule
			NRMSE & \textbf{1.82e-2} & 1.83e-2 & 1.87e-2 & 1.87e-2 & 1.86e-2 & 1.95e-2 \\
			NLL   & -3.34  & -3.97  & \textbf{-4.19}  & -3.27 & -4.06 & -4.18 \\
			\bottomrule
		\end{tabular}
	\end{subtable}
	\vspace{0.1in}
	
	\begin{subtable}{\textwidth}
		\centering
		\caption{\small Choice of kernels.}
		\begin{tabular}{ccccccc}
			\toprule
			\textbf{GP Kernel} & Squared Exp & Matérn 13/2 & Matérn 5/2 & \multicolumn{3}{c}{Weighted Matérn (5/2+13/2)} \\
			\midrule
			NRMSE & 2.78e-2 & 1.92e-2 & 2.05e-2 & \multicolumn{3}{c}{\textbf{1.86e-2}} \\
			NLL   & -2.98  & -3.43  & -3.52  & \multicolumn{3}{c}{\textbf{-4.06}} \\
			\bottomrule
		\end{tabular}
	\end{subtable}
\end{table}

\noindent\textbf{Projection Points.} Next, we examined the effect of the number of projection points, \ie the quadrature nodes or sampling locations used across GP layers. We performed an ablation study on the  \textit{Burgers} dataset with varying numbers of projection points, using Gauss-Legendre quadrature. The quadrature resolution directly determines the number of the weight function values to estimate and thus the number of trainable parameters.

As shown in Table~\ref{tab:quad-number}, model performance degrades when the number of projection points is too small (\eg 8 or 16). However, beyond a certain threshold (64), additional points provide little to no improvement while substantially increasing the parameter count and computational cost.
\begin{table}[H]
	\centering
	\small
	\caption{\small Performance of DGPFM-QR with Gauss-Legendre quadrature on \textit{Burgers} dataset.}\label{tab:quad-number}
	\begin{tabular}{lcccccc}
		\toprule
		\#Project Points & 8 & 16 & 32 & 64 & 128 & 256 \\
		\midrule
		 NRMSE (\%) & 13.684 & 2.755 & 0.982 & \textbf{0.676} & 0.796 & 0.892 \\
		\bottomrule
	\end{tabular}
\end{table}
\noindent\textbf{GP Activation.} To assess the benefit of our GP-based activation, we conducted ablation studies comparing it against standard non-probabilistic activations commonly used in neural networks (ReLU and Tanh), as well as against the case with no nonlinear activation.

As shown in Table~\ref{tab:activation}, the GP-based activation substantially improves both training and test errors relative to ReLU and Tanh (by more than 10\%), with the sole exception of \textit{Beijing-Air}, where the test error is marginally higher (a relative increase of 1.8\%). Removing the nonlinear activation entirely results in a large increase in test error, underscoring its critical role in model performance.

\begin{table}[t]
	\small
	\centering
	\caption{\small \ours with different activate functions. The base models are the same in Table~\ref{tab:dgpfm-ft-ablation} and~\ref{tab:dgpfm-qr-ablation}}\label{tab:activation}
	\begin{subtable}[t]{\textwidth}
		\centering
		\caption{\small DGPFM-FT on \textit{Beijing Air}. }
		\begin{tabular}{lcccc}
			\toprule
			\textbf{Activation} & ReLU & Tanh & GP Activation & No Activation \\
			\midrule
			Training NRMSE(\%) & 0.6916 & 0.6953 & \textbf{0.028901} & 53.275 \\
			Test NRMSE(\%)  & \textbf{26.363} & 37.402 & {26.854}   & 57.615 \\
			\bottomrule
		\end{tabular}
	\end{subtable}%
	\vspace{0.1in}
	\begin{subtable}[t]{\textwidth}
		\centering
		\caption{\small DGPFM-QR on the \textit{Darcy} dataset. }
		\begin{tabular}{lcccc}
			\toprule
			\textbf{Activation} & ReLU & Tanh & GP Activation & No Activation \\
			\midrule
			Training NRMSE (\%) & 1.581 & 1.737 & \textbf{1.247} & 5.395 \\
			Test NRMSE (\%)  & 2.131 & 2.0292 & \textbf{1.824} & 6.703 \\
			\bottomrule
		\end{tabular}
	\end{subtable}
\end{table}

 \begin{table}[H]
 \centering
 \small
 \caption{\small Performance of \ours-QR with different integral transforms on the \textit{Darcy} dataset.}\label{tab:dimwise-ablation}
 \begin{tabular}{ccc}
 \toprule
  & {Dimension-wise} & {Full-dimensional} \\
 \midrule
 Training NRMSE (\%) & 1.247 & \textbf{0.0830} \\
 Test  NRMSE (\%)  & \textbf{1.824} & 3.669 \\
 \bottomrule
 \end{tabular}
 \end{table}

\noindent\textbf{Dimension-Wise and Full Integral Transform.} To evaluate the effectiveness of our dimension-wise discrete integral transform, we trained a DGPFM-QR model on \textit{Darcy} with and without the dimension-wise transform, using five integration layers, 64 channels, 64 inducing points, and a weighted Matérn kernel (DOF 5/2 and 13/2). The trapezoidal rule with 29 projection points per input dimension was employed. As shown in Table~\ref{tab:dimwise-ablation}, the full-dimensional transform fits the training data more closely but performs substantially worse on the test set, indicating clear overfitting.
\begin{figure}[H]
	\centering
	\includegraphics[angle=270, width=\linewidth]{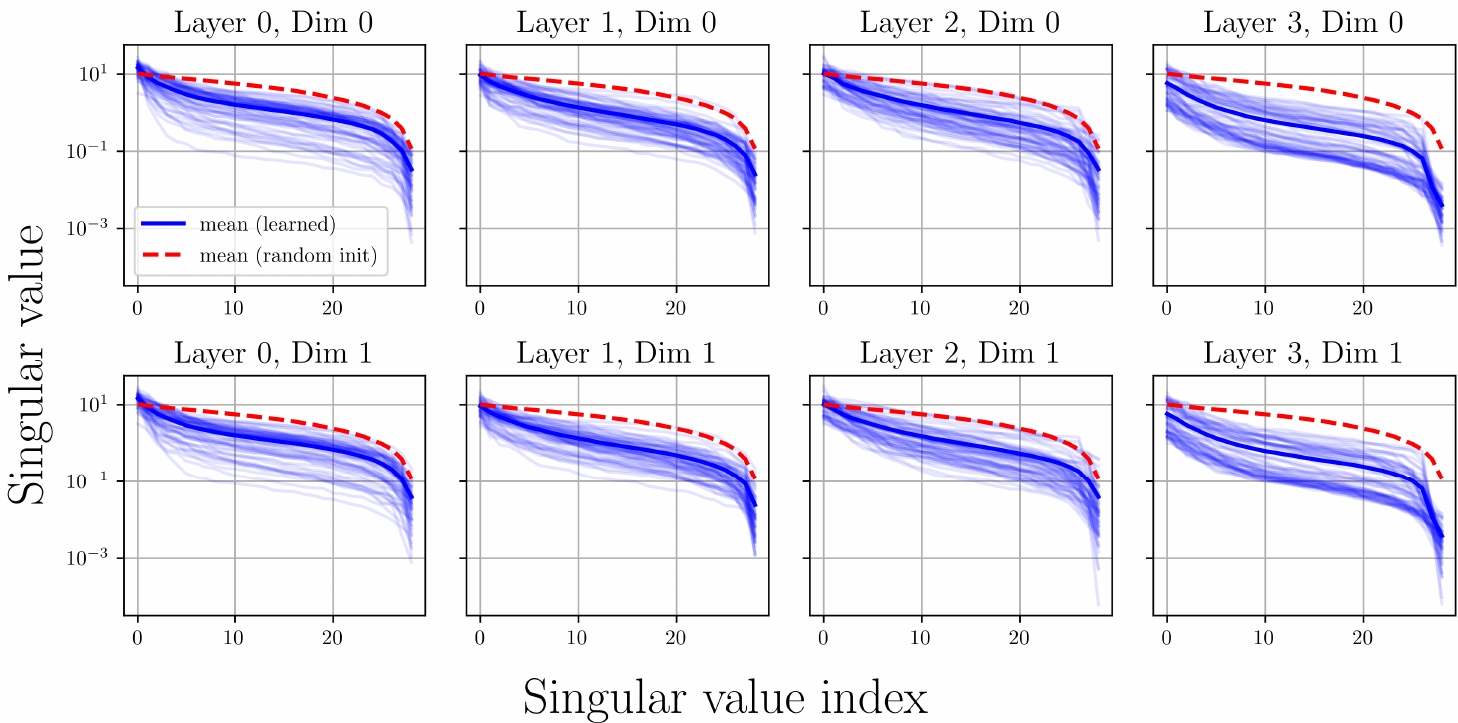}
	\caption{\small The Singular values of learned weight function values (matrices) versus randomly initialized matrices for running \ours-QR on the \textit{Darcy} dataset.}
	\label{fig:darcy_singular_values}
\end{figure}
%\begin{figure}[H]
%	\centering
%	\includegraphics[angle=270, width=1.\linewidth]{figs/darcy_examples_pair_3-trim.pdf}
%	\caption{Plot of solution functions for DGPFM-QR on the Darcy-Flow problem.}
%	\label{fig:darcy_plot}
%\end{figure}

\section{Visualization of Learned Weight Functions}\label{sect:weight}
To better understand the representations learned by the weight functions, we applied \ours-QR to the \textit{Darcy} dataset and analyzed the weight matrices in the integration (linear) layers. Specifically, we performed Singular Value Decomposition (SVD) on the weight matrix associated with each input dimension. As shown in Figure~\ref{fig:darcy_singular_values}, in the early layers the singular values decay slowly, indicating that the weight matrices remain close to full rank and thus capture diverse, information-rich features. In contrast, in the layers closer to the output, the singular values decay more rapidly, suggesting the emergence of low-rank structures. This progression highlights a representational shift: early layers emphasize broad feature extraction, while later layers distill these features into more compact and structured representations tailored for prediction.

\section{Running Time}\label{sect:time}
To evaluate training efficiency, we compared our method with the neural operator models GNOT and FNO on the \textit{Darcy} and \textit{Car Shape} datasets. All experiments were conducted under the same computing environment --- a Linux workstation equipped with an NVIDIA GeForce RTX 4080 GPU. We report both the training time per epoch and the total training time (i.e., until the stopping criterion is met). The results are summarized in Table~\ref{tab:time}.

On the \textit{Darcy} dataset, \ours (both \ours-FT and \ours-QR) and FNO exhibit comparable performance in terms of both per-epoch runtime and overall training time. On the \textit{Car Shape} dataset, DFT-FNO is the fastest, as it directly applies NUDFT without requiring interpolation of function samples onto a regular grid. \ours-QR achieves the second-fastest training time, while \ours-FT is comparable to GNOT. The additional computational cost of \ours-FT arises from the conditional GP, which interpolates irregularly sampled input functions onto a regular grid to enable standard Fourier transforms. This overhead could be reduced by directly incorporating NUDFT into our integral transform, which we leave for future work.

\begin{table}[H]
	\centering
	\small
	\caption{\small Training time comparison. All the methods were run on a Linux workstation with a NVIDIA GeForce RTX 4080 GPU.  \cmt{Collected on a NVIDIA GeForce RTX 4080 assuming batch sizes of 1 due to memory constraints.}}\label{tab:time}
	\begin{subtable}{\textwidth}
		\centering
		\caption{\small \textit{Darcy}}
		\begin{tabular}{lcc}
			\toprule
			\textbf{Method} & Per-epoch (seconds) & Total (min) \\
			\midrule
			GNOT         & 0.133 & 220.844 \\
			FNO         & 0.00945 & 15.750\\
			%DGPFM-FT         & 0.0214 & 35.638\\
			%DGPFM-QR         & 0.0167 & 27.818 \\
			\ours-FT        & 0.0241 & 40.239\\
			\ours-QR      &  0.0184 & 30.622\\
			\bottomrule
		\end{tabular}
	\end{subtable}
	\begin{subtable}{\textwidth}
		\centering
		\caption{\small \textit{Car Shape}}
		\begin{tabular}{lcc}
			\toprule
			\textbf{Method} & Per-epoch (seconds) / Step & Total (min) \\
			\midrule
			GNOT         & 9.69 & 48.50 \\
			DFTFNO        & 0.0883 & 1.47 \\
			%DGPFM-FT         &  0.0213 & 319.423  \\
			%DGPFM-QR         &   0.0181 & 272.091\\
			\ours-FT        &  11.15 & 95.83  \\
			\ours-QR      &  0.933 &  15.55 \\
			\bottomrule
		\end{tabular}
	\end{subtable}
%		\begin{subtable}{\textwidth}
%		\centering
%		\caption{\small \textit{3D NS}}
%		\begin{tabular}{lcc}
%			\toprule
%			\textbf{Method} & Per-epoch (seconds) / Step & Total (min) \\
%			\midrule
%			GNOT         & 0.2213 & 3319.611\\
%			FNO         & 0.0383 & 570.375 \\
%			%DGPFM-FT         &  0.0213 & 319.423  \\
%			%DGPFM-QR         &   0.0181 & 272.091\\
%			\ours-FT        &  0.0218 & 326.947  \\
%			\ours-QR      &  0.0221 & 331.990  \\
%			\bottomrule
%		\end{tabular}
%	\end{subtable}
	
\end{table}
%
%\begin{table}[H]
%	\centering
%	\caption{Training time on the \textit{3D NS}. \cmt{Collected on a NVIDIA GeForce RTX 4080 assuming batch sizes of 1 due to memory constraints.}}
%	\begin{tabular}{lcc}
%		\toprule
%		\textbf{Method} & Seconds / Step & Total (min) \\
%		\midrule
%		GNOT         & 0.2213 & 3319.611\\
%		FNO         & 0.0383 & 570.375 \\
%		DGPFM-FT         &  0.0213 & 319.423  \\
%		DGPFM-QR         &   0.0181 & 272.091\\
%		DGPFM-FT SVI        &  0.0218 & 326.947  \\
%		DGPFM-QR SVI      &  0.0221 & 331.990  \\
%		\bottomrule
%	\end{tabular}
%\end{table}
%
%
%\begin{table}[H]
%	\centering
%	\caption{Training time comparison on the 2D \emph{Darcy-Flow} dataset. Collected on a NVIDIA GeForce RTX 4080 assuming batch sizes of 100.}
%	\begin{tabular}{lcc}
%		\toprule
%		\textbf{Method} & Seconds / Step & Total (min) \\
%		\midrule
%		GNOT         & 0.133 & 220.844 \\
%		FNO         & 0.00945 & 15.750\\
%		DGPFM-FT         & 0.0214 & 35.638\\
%		DGPFM-QR         & 0.0167 & 27.818 \\
%		DGPFM-FT SVI        & 0.0241 & 40.239\\
%		DGPFM-QR SVI      &  0.0184 & 30.622\\
%		\bottomrule
%	\end{tabular}
%\end{table}

\section{Limitation}
Our current designs of the discrete integral transform are relatively simple and may be limited in scope. The associated weight functions \cmt{--- equipped with GP priors ---} are global over the domain and thus may struggle to capture local, nonstationary patterns. In future work, we aim to explore richer classes of transforms, such as spline-based or orthogonal basis functions, as well as localized sparse functions. These alternatives could further reduce model complexity, improve expressiveness, and provide finer control over the granularity of integration, thereby enabling the model to adapt more effectively to heterogeneous or highly localized structures in functional data.
%Our current designs of the discrete integral transform are simple but might be limited in scope. These weight functions --- assigned with GP prior --- are global in the domain, and might not be able to capture the local, nonstationary patterns.  In future work, we plan to explore more transforms --- such as those based on splines or orthogonal basis
%functions, and local sparse functions --- to further reduce model complexity and better control integration granularity. 
%\input{./appendix-new}

\end{document}